\documentclass[letterpaper]{article}
\usepackage[preprint]{aaai2027}

\usepackage[hyphens]{url}
\usepackage{graphicx}
\usepackage{natbib}
\usepackage{caption}
\usepackage{flafter}
\usepackage{booktabs}
\usepackage{multirow}
\usepackage[table]{xcolor}
\usepackage{colortbl}
\usepackage{xspace}
\usepackage{subcaption}
\usepackage{pifont}
\usepackage{marvosym}
\usepackage{makecell}
\usepackage{amsmath}
\usepackage{amsfonts}
\usepackage{dsfont}
\usepackage{bm}

\usepackage{amsmath,amsfonts,bm}

\def\eqref#1{equation~\ref{#1}}

\def\1{\bm{1}}

\DeclareMathAlphabet{\mathsfit}{\encodingdefault}{\sfdefault}{m}{sl}
\SetMathAlphabet{\mathsfit}{bold}{\encodingdefault}{\sfdefault}{bx}{n}

\newcommand{\arxivcodeavailability}{Code is available at \url{https://github.com/YangYangGirl/ParVL}.}

\definecolor{navyblue}{HTML}{0071BC}

\title{ParVL: Parallel Scaling and Expandable \\Compute Allocation for Multimodal LLMs}

\author{
    Yang Yang\textsuperscript{\rm 1,\rm 2,}\thanks{Corresponding author: yang.yang3@anu.edu.au},
    Qinyu Zhao\textsuperscript{\rm 1},
    Mouxiang Chen\textsuperscript{\rm 3},
    Xiaohui Li\textsuperscript{\rm 4},\\
    Lixin Gu\textsuperscript{\rm 2},
    Wenhai Wang\textsuperscript{\rm 5},
    Hongjie Zhang\textsuperscript{\rm 2},
    Wenwei Zhang\textsuperscript{\rm 2}
}
\affiliations{
    \textsuperscript{\rm 1}The Australian National University \quad
    \textsuperscript{\rm 2}Shanghai AI Laboratory\\
    \textsuperscript{\rm 3}Zhejiang University \quad
    \textsuperscript{\rm 4}Shanghai Jiao Tong University \quad
    \textsuperscript{\rm 5}Nanjing University
}

\begin{document}

\maketitle

\begin{abstract}

Existing scaling strategies for Multimodal Large Language Models (MLLMs) typically expand either model parameters or sequential inference computation, incurring substantial memory or latency overhead. More importantly, most existing methods fail to alter the rigid, fixed computation allocation between the Vision Transformer and the Large Language Model components, limiting task-specific optimization. To address this, we introduce the Parallel Vision-Language (ParVL) scaling framework for MLLMs, which scales parallel computation by reusing the existing ViT and LLM backbone parameters across multiple vision and language branches. This framework raises a central question: given a fixed backbone parameter budget, how should additional shared-backbone computation be allocated between the vision and language modalities? We instantiate each parallel computational stream with branch-specific prefix parameters over a shared backbone, and train the entire model end-to-end via full-parameter supervised fine-tuning on roughly 13B tokens. We systematically study the computation-allocation trade-off between the ViT encoder and LLM decoder. ParVL improves overall multimodal performance over same-recipe single-branch baselines, and the best evaluated vision--language allocation varies across tasks.\ifdefined\arxivcodeavailability\space\arxivcodeavailability\fi

\end{abstract}

\begin{figure}[t]
\centering
\includegraphics[width=0.93\linewidth]{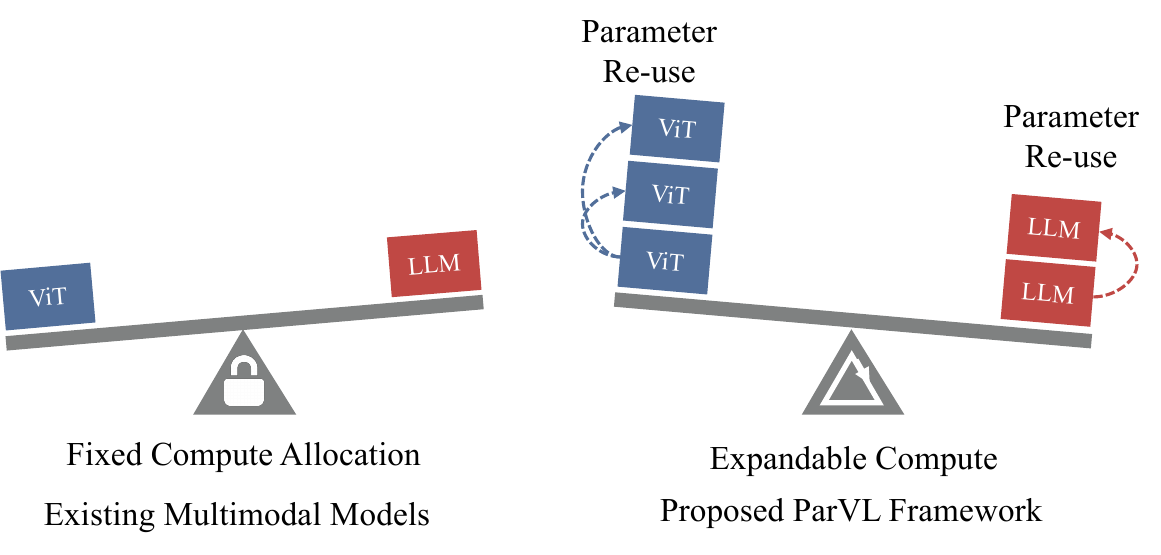}
\caption{\textbf{Fixed vs. expandable compute allocation in MLLMs.}
\textbf{(Left) Existing multimodal models} suffer from a fixed compute allocation between ViT (vision) and LLM (language) components during the SFT phase, which is visualized as a locked seesaw.
\textbf{(Right) Our ParVL framework} enables expandable and flexible compute for the two modules. By leveraging extensive parameter reuse, our parallel architecture independently scales computation in the vision and language modules during SFT, enabling controlled comparison of different vision--language allocations across branch-count configurations.}
\label{fig:motivation}
\end{figure}

\section{Introduction}
\label{sec:intro}

Large-model capability is commonly scaled by increasing either parameter count or computation. Increasing parameters raises storage and memory costs, while many computation-scaling approaches, including Chain-of-Thought (CoT) reasoning~\citep{cot} and iterative refinement~\citep{madaan2023self}, add sequential inference steps and therefore increase response latency. Parameter reuse provides a complementary direction: shared parameters can be applied recurrently or across parallel streams to increase computation while keeping the backbone parameter count fixed. Recent examples include LoopLM~\citep{zhu2025scaling}, ParScale~\citep{chen2025parallel}, and Mixture-of-Recursions (MoR)~\citep{NEURIPS2025_8b08bbf8}. These developments motivate parameter-efficient computation scaling under a fixed backbone parameter budget.

However, extending parameter-reuse scaling to MLLMs introduces an additional allocation problem: extra computation can be assigned to either the ViT encoder or the LLM decoder. High-resolution document understanding and fine-grained grounding can place greater demands on visual processing, whereas long-form reasoning can stress the decoder; many tasks require both, but not in the same proportion.

Most conventional MLLMs determine this balance at design time rather than expose vision and language computation as independently controllable scaling axes, as illustrated in Figure~\ref{fig:motivation} (left). This raises an MLLM-specific question: given a fixed backbone parameter budget, how can computation be expanded through parameter reuse while controlling its allocation between vision and language? Our experiments show that the preferred ViT/LLM allocation varies across benchmarks with different visual and reasoning demands.

In this work, we introduce Parallel Vision-Language (ParVL) scaling, a parameter-sharing framework that formulates parallel scaling in MLLMs as a two-dimensional vision-language compute-allocation problem. Because these streams can execute concurrently, additional computation need not translate into proportional wall-clock latency growth when hardware resources permit. To the best of our knowledge, ParVL is the first study to jointly scale the ViT encoder and LLM decoder through parallel parameter reuse. ParVL constructs separate prefix-conditioned branches for the two components, aggregates their representations at modality-specific stages, and controls their branch counts independently. This design enables expanded computation to be allocated flexibly between vision and language with modest parameter growth.

Under the same full-parameter SFT recipe, the best evaluated ParVL configuration improves the 9-benchmark average from 49.6 to 50.5 at 1B. At the 2B and 8B scales, the balanced branch-count configuration ($P_v = 2, P_l = 2$) yields average scores of 54.7 and 63.0, compared with 54.4 and 62.5 for the corresponding single-branch baselines.

Our contributions are threefold. (1) We introduce ParVL, a parallel-scaling framework for MLLMs that formulates computation expansion as a two-dimensional vision--language allocation problem and implements it with parameter-shared, prefix-conditioned ViT and LLM branches and modality-specific token-wise aggregation. (2) Through a systematic study of nine vision--language branch-count configurations, we show that the preferred compute allocation varies across benchmarks. (3) Controlled comparisons at 1B, 2B, and 8B scales, together with latency and memory profiling, characterize the accuracy--cost trade-offs of parallel scaling.

\section{Related Work}
\label{sec:related_work}

\paragraph{Scaling Laws for Language Models.}
Scaling laws characterize predictable relationships between model performance and parameter count, training-data size, and training compute~\citep{kaplan2020scaling,hoffmann2022training}. Dense parameter scaling increases capacity but also raises storage and memory costs. Sparse MoE models instead expand total capacity while activating only a subset of experts for each token~\citep{jacobs1991adaptive,shazeer2017outrageously,fedus2022switch,du2022glam,tian2025towards}. Computation scaling instead allocates additional computation without necessarily enlarging the parameter set. A complementary scaling direction is to reuse model parameters across recurrent or parallel computation paths. Recent recursive and recurrent examples include LoopLM~\citep{zhu2025scaling}, Mixture-of-Recursions (MoR)~\citep{NEURIPS2025_8b08bbf8}, and Relaxed Recursive Transformers~\citep{bae2025relaxed}. LoopLM performs iterative latent computation by repeatedly applying a shared model, MoR reuses a shared layer stack across dynamically assigned recursion depths, and Relaxed Recursive Transformers pair shared base layers with layer-wise LoRA adapters. These methods scale effective depth through sequential parameter reuse. ParScale explores a parallel alternative by reusing a shared model across multiple concurrent streams and aggregating their outputs~\citep{chen2025parallel}.

Sparse-capacity scaling has also been adopted in MLLMs. MoE-LLaVA replaces selected LLM feed-forward layers with sparse expert layers, whereas CuMo places sparsely gated MoE blocks in both the vision encoder and MLP connector~\citep{lin2026moe,li2024cumo}. These methods add and route expert parameters rather than reusing a shared dense backbone across repeated computation. In contrast, shared-backbone parameter reuse in MLLMs remains largely unexplored. The extension is non-trivial as it introduces a critical new scientific question that does not exist in the text-only domain: given a fixed parameter budget, how can one effectively expand the model's computational capacity via parameter reuse, and concurrently determine how computational resources should be allocated between the vision (ViT) and language (LLM) modalities? In this work, we are the first to systematically investigate this vision-language computation allocation trade-off. We adapt a parallel architecture inspired by ParScale merely as a \textit{vehicle} for this study, proposing our ParVL framework to flexibly control and evaluate this novel resource allocation problem.


\paragraph{Inference-Time Scaling.}
The success of reasoning-oriented language models, exemplified by DeepSeek-R1~\citep{guo2025deepseek}, has intensified interest in inference-time scaling, which allocates additional computation during inference to improve model performance. Serial inference-time scaling commonly allocates additional computation through longer Chain-of-Thought reasoning traces~\citep{cot}, thereby extending the serial inference path and typically increasing latency. Instead, parallel scaling expands the inference computation across multiple candidate paths. The beam search maintains multiple partial hypotheses during decoding~\citep{wiseman2016sequence}, whereas Self-Consistency samples diverse Chain-of-Thought trajectories and aggregates their final answers~\citep{wang2022self}. Where hardware resources permit, candidate paths can be evaluated concurrently, so increasing inference computation does not need to translate linearly into wall-clock latency. 

However, concurrency does not eliminate redundancy across independently generated paths, motivating methods that coordinate parallel reasoning or prune redundant trajectories. Adaptive parallel reasoning learns to invoke \texttt{spawn()} and \texttt{join()} operations to coordinate serial and parallel reasoning threads~\citep{pan2025learning}. ThreadWeaver trains language models to organize reasoning into adaptive threads and executes them through a trie-based design compatible with autoregressive inference engines~\citep{lian2025threadweaver}. DeepPrune detects equivalent partial reasoning trajectories and prunes redundant paths before completion~\citep{tu2026deepprune}. These methods improve the efficiency of output-space parallel scaling by structuring or pruning concurrent reasoning paths. ParVL applies parallel scaling to internal computation streams within an MLLM rather than to output trajectories. The challenge is not simply to generate more language candidates, but to determine how shared-backbone parallel computation should be divided between vision and language to balance performance, latency, and memory.

\section{Method}
\label{sec:method}

\begin{figure*}[t]
\centering
\includegraphics[width=\linewidth]{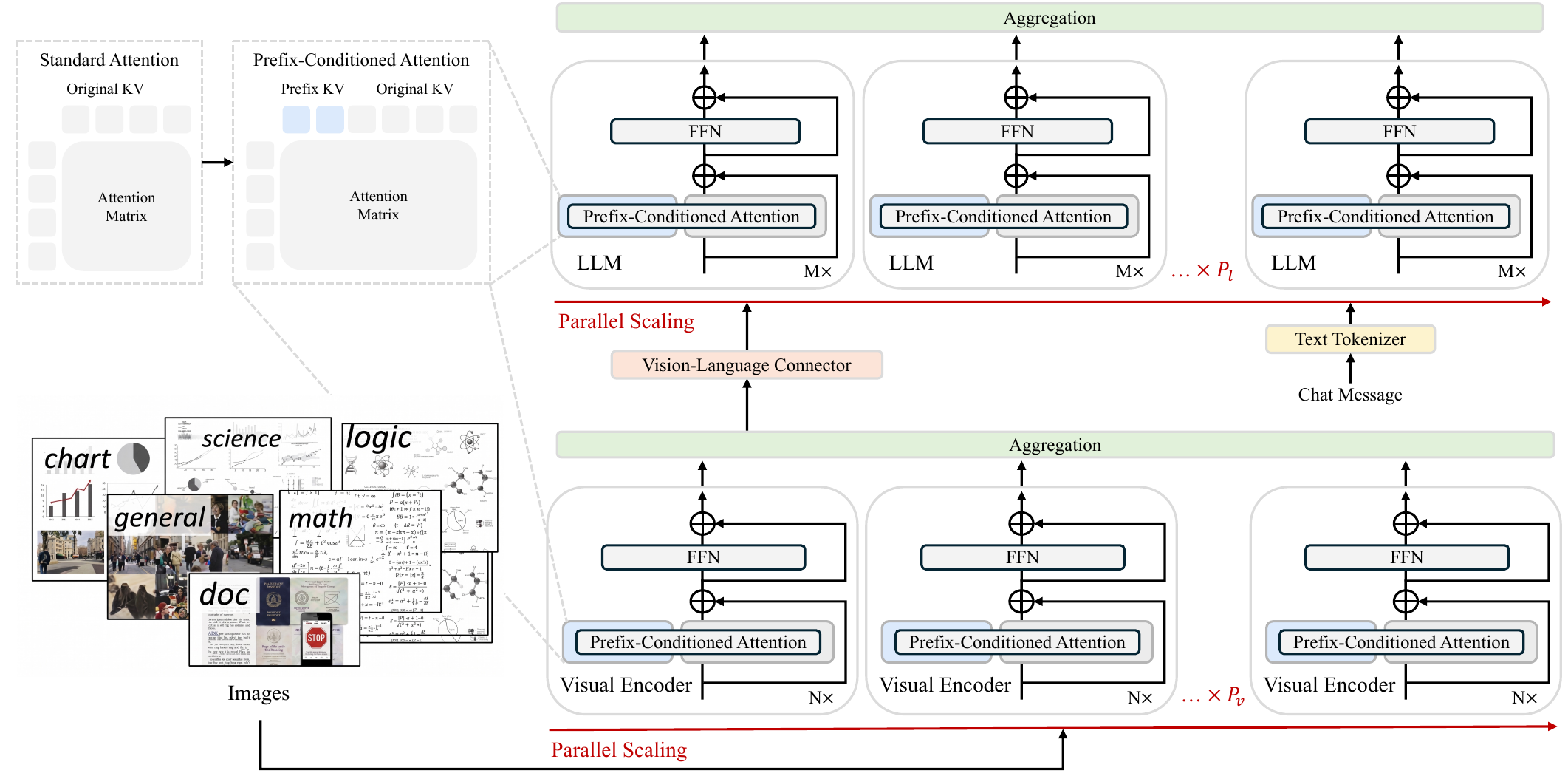}
\caption{\textbf{Overall architecture.} ParVL scales multimodal computation with $P_v$ parameter-shared ViT branches and $P_l$ parameter-shared LLM branches in parallel. Learnable KV prefixes differentiate branches within each shared backbone. A token-wise MLP aggregator fuses vision states before the connector, whose projected tokens are replicated as common input to all language branches. A separate token-wise MLP aggregator fuses language states before the shared language-modeling head. All components are jointly optimized by full-parameter SFT.}
\label{fig:architecture}
\end{figure*}

\paragraph{Architecture Overview.}

ParVL scales vision and language computation with $P_v$ ViT branches and $P_l$ LLM branches (Figure~\ref{fig:architecture}) in parallel. We denote a branch-count configuration by $P_v{:}P_l$, listing the number of ViT branches first and the number of LLM branches second. Branches within each component share all backbone weights and are differentiated only by branch-specific key and value prefixes introduced at each prefix-conditioned attention layer. Adding a branch therefore increases computation with modest parameter overhead. Vision outputs are aggregated and projected into the LLM embedding space, replicated across language branches, and fused again before the shared language-modeling head.



\paragraph{Prefix-Conditioned Attention.}
\label{sec:prefix_conditioned_attention}
For clarity, we describe a single self-attention head and omit the layer index. Given a hidden sequence $X\in\mathbb{R}^{L\times d}$, standard attention computes $Q=XW_Q$, $K=XW_K$, and $V=XW_V$, where $Q,K,V\in\mathbb{R}^{L\times d_h}$. Branch $k$ maintains learnable prefixes $K_P^{(k)},V_P^{(k)}\in\mathbb{R}^{L_p\times d_h}$ while sharing $W_Q,W_K,W_V$ with all other branches. We prepend the prefixes along the sequence dimension,
\begin{equation}
\begin{aligned}
\tilde K^{(k)}&=[K_P^{(k)};K],\qquad
\tilde V^{(k)}=[V_P^{(k)};V],\\
O^{(k)}
&=
\operatorname{softmax}\!\left(
\frac{Q^{(k)}\tilde K^{(k)\top}}{\sqrt{d_h}}
+
M
\right)\tilde V^{(k)},
\end{aligned}
\label{eq:prefix_attention}
\end{equation}
where $M\in\mathbb{R}^{L\times(L_p+L)}$ is an all-zero mask for bidirectional ViT attention and a causal mask for the LLM.
Prefixes are prepended only to the key and value sequences; the query sequence remains unprefixed. During autoregressive decoding, the learned prefix keys and values are pre-seeded in each branch's KV cache.

Prefixes are instantiated in every self-attention layer and are the only branch-specific backbone parameters. Attention still returns states for the original $L$ positions; prefix positions serve only as learned context. Their parameter cost grows with prefix length and branch count, but remains small relative to duplicating a ViT or LLM backbone. During language-side decoding, caching each branch's learned prefix makes its specialization available from the first decoding step without modifying the shared projection matrices.



\paragraph{Vision and Language Branch Aggregation.}
\label{sec:aggregation_training}
Let $Z_{m,k}\in\mathbb{R}^{L_m\times D_m}$ denote the hidden states from branch $k$ of modality $m\in\{v,\ell\}$, where $P_m$ is the corresponding branch count and $z_{m,k,t}\in\mathbb{R}^{D_m}$ is the feature at token position $t$. For each token, we concatenate the branch features as $c_{m,t}=\operatorname{Concat}(z_{m,1,t},\ldots,z_{m,P_m,t})\in\mathbb{R}^{P_mD_m}$. A modality-specific two-layer SiLU MLP $g_m:\mathbb{R}^{P_mD_m}\rightarrow\mathbb{R}^{P_m}$ then predicts the branch logits:
\begin{equation}
\begin{aligned}
a_{m,t}
&=\operatorname{softmax}\!\left(g_m(c_{m,t})\right),\\
\bar a_{m,t,k}
&=(1-\lambda_m)a_{m,t,k}+\frac{\lambda_m}{P_m},\\
z_{m,t}
&=\sum_{k=1}^{P_m}\bar a_{m,t,k}z_{m,k,t}.
\end{aligned}
\label{eq:modality_agg}
\end{equation}
The softmax operates over branches, and the aggregated sequence is $Z_m=(z_{m,t})_{t=1}^{L_m}$. The smoothing coefficient $\lambda_m\in[0,1]$ interpolates the predicted weights with a uniform distribution, discouraging premature concentration on only a few branches when $\lambda_m>0$. All main experiments use MLP aggregation, with $\lambda_v=0$ and $\lambda_\ell=0.1$. The supplement provides a comparison with mean aggregation.

For vision, $Z_{v,k}$ comes from prefix-conditioned InternViT branches. After aggregation, we remove the class token:
\begin{equation}
Z_v^{\mathrm{patch}}
=
\operatorname{RemoveCLS}(Z_v)
\in\mathbb{R}^{(L_v-1)\times D_v}.
\end{equation}

We then reshape the patch sequence into its spatial grid and apply the InternVL pixel-shuffle operation:
\begin{equation}
\widehat Z_v
=
\Pi_r(Z_v^{\mathrm{patch}})
\in\mathbb{R}^{N\times D_c},
\end{equation}
where $r$ is the pixel-shuffle ratio, $N$ is the resulting number of
visual tokens, and $D_c$ is their channel dimension. The InternVL
connector maps these tokens to the LLM hidden space:
\begin{equation}
H_v=W_2\operatorname{GELU}\!\left(W_1\operatorname{LN}(\widehat Z_v)\right).
\end{equation}
Here, the connector maps the aggregated visual tokens to $H_v\in\mathbb{R}^{N\times D_\ell}$. Each language branch receives the same $H_v$ and text embeddings, processes them with the shared LLM backbone under a branch-specific prefix, and produces a hidden-state sequence. Equation~\ref{eq:modality_agg} fuses these language states before the shared language-modeling head produces next-token logits.

\paragraph{Training and Compute Allocation.}
We jointly train the shared backbones, branch prefixes, aggregation modules, and connector using a weighted next-token prediction loss over valid response tokens. Let $b$ index training sequences, $t$ index token positions, and $S_{q,b}$ denote the prompt context for sequence $b$. For sequence $b$, define $n_b=\sum_t m_{b,t}$ and $\omega_b=1/\sqrt{n_b}$, where $m_{b,t}=1$ for valid target tokens and $m_{b,t}=0$ for prompt or padding tokens. The SFT objective is
\begin{equation}
\mathcal L_{\mathrm{SFT}}
=
-\frac{
\sum_b\sum_t m_{b,t}\omega_b
\log p_\theta(y_{b,t}\mid y_{b,<t},H_{v,b},S_{q,b})
}{
\sum_b\sum_t m_{b,t}\omega_b
}.
\label{eq:loss}
\end{equation}
The denominator normalizes the weighted loss over all valid target tokens in the batch. Here, the normalization with $\omega_b=1/\sqrt{n_b}$ reduces the dominance of long responses while preserving their greater amount of supervision.


ParVL uses full-parameter SFT, so gradients from all active branches update the shared
backbone parameters, while the prefixes and aggregation modules learn branch-specific conditioning and fusion. We vary $P_v$ and $P_l$ over $\{1,2,4\}$, covering balanced, vision-heavy, and language-heavy branch-count configurations. This allocation study addresses two questions: whether parallel reuse improves the same-recipe single-branch baseline, and how the extra computation should be allocated between the vision and language components for each task.

Because the branch-count configuration ($P_v = 4, P_l = 4$) entails more computation than $1{:}1$, the accuracy difference between these configurations reflects both compute-scale and vision--language allocation effects. We therefore compare against the same-recipe single-branch baseline to assess expanded computation and report the FLOPs of each configuration when analyzing allocation. Since vision and language branches have different computational costs, these comparisons characterize empirical accuracy--cost trade-offs rather than isolate an allocation effect under exactly matched compute. Parameter and FLOP estimates for representative settings are reported in the supplement; across the evaluated configurations, total parameters increase by at most 4\% over the corresponding single-branch baselines.

\section{Experiments}
\label{sec:experiment}

\begin{table*}[t!]
\centering
\fontsize{8}{10}\selectfont
\setlength{\tabcolsep}{1.1mm}
\renewcommand{\arraystretch}{1.12}

\resizebox{\linewidth}{!}{
\begin{tabular}{l|ccccccccc|cccc}
\toprule
\multirow{2}{*}{\textbf{Model}}
 & \multicolumn{1}{c}{\textbf{General}} & \multicolumn{4}{c}{\textbf{Math}} & \multicolumn{4}{c|}{\textbf{OCR}} & \multicolumn{4}{c}{\textbf{Average}} \\
\cmidrule(lr){2-2} \cmidrule(lr){3-6} \cmidrule(lr){7-10} \cmidrule(lr){11-14}
 & MMMU$_{dev\text{-}val}$ & MathVista$_{mini}$ & MathVision$_{mini}$ & LogicVista & WeMath
 & ChartQA$_{test}$ & TextVQA$_{val}$ & DocVQA$_{val}$ & OCRBench
 & General & Math & OCR & \textbf{All} \\
\midrule
\multicolumn{14}{l}{\textit{$\sim$\,1B Models}} \\
\midrule
InternVL3.5-1B-Pretrained~\citep{wang2025internvl3} & 38.0 & 38.5 & 16.8 & 27.3 & 12.9 & 71.0 & 67.8 & 77.3 & 70.5 & 38.0 & 23.9 & 71.7 & 46.7 \\
InternVL3.5-1B-Instruct~\citep{wang2025internvl3} & 39.9 & 50.7 & 22.0 & 32.0 & 12.5 & 77.2 & 71.1 & 84.0 & 77.9 & 39.9 & 29.3 & 77.6 & 51.9 \\
\midrule
\rowcolor{blue!8}Baseline ($P_v = 1, P_l = 1$) & 40.3 & 43.9 & 21.1 & 27.3 & 11.6 & 71.7 & 71.4 & 82.6 & 76.6 & \textbf{40.3} & 26.0 & 75.6 & 49.6 \\
\rowcolor{blue!8}ParVL-1B ($P_v = 4, P_l = 1$) & 39.7 & 45.0 & 22.4 & 30.0 & 11.3 & 72.9 & 71.5 & 82.2 & 75.5 & 39.7 & 27.2 & 75.5 & 50.1 \\
\rowcolor{blue!8}ParVL-1B ($P_v = 1, P_l = 4$) & 39.8 & 44.5 & 24.7 & 29.5 & 11.6 & 72.1 & 71.6 & 82.4 & 76.5 & 39.8 & 27.6 & 75.7 & 50.3 \\
\rowcolor{blue!8}ParVL-1B ($P_v = 4, P_l = 4$) & 39.0 & 48.0 & 22.4 & 29.3 & 12.3 & 72.1 & 71.9 & 82.7 & 77.0 & 39.0 & \textbf{28.0} & \textbf{75.9} & \textbf{50.5} \\
\midrule
\multicolumn{14}{l}{\textit{$\sim$\,2B Models}} \\
\midrule
Ovis2-2B~\citep{lu2024ovis} & 42.6 & 64.5 & 19.4 & 33.6 & 10.3 & 81.3 & 79.9 & 91.6 & 87.1 & 42.6 & 32.0 & 85.0 & 56.7 \\
Qwen3-VL-2B-Instruct~\citep{bai2025qwen3} & 44.4 & 54.2 & 20.7 & 31.8 & 28.4 & 77.7 & 80.8 & 93.0 & 86.8 & 44.4 & 33.8 & 84.6 & 57.5 \\
InternVL3.5-2B-Pretrained~\citep{wang2025internvl3} & 45.7 & 50.3 & 17.4 & 30.9 & 16.3 & 77.8 & 73.6 & 85.4 & 77.7 & 45.7 & 28.7 & 78.6 & 52.8 \\
InternVL3.5-2B-Instruct~\citep{wang2025internvl3} & 49.9 & 61.4 & 25.0 & 40.5 & 19.9 & 80.6 & 76.1 & 88.3 & 82.9 & 49.9 & 36.7 & 82.0 & 58.3 \\
\midrule
\rowcolor{blue!8}Baseline ($P_v = 1, P_l = 1$) & 46.4 & 49.3 & 19.7 & 33.6 & 18.4 & 76.7 & 76.0 & 88.3 & 81.5 & \textbf{46.4} & 30.3 & 80.6 & 54.4 \\
\rowcolor{blue!8}ParVL-2B ($P_v = 2, P_l = 2$) & 46.1 & 53.6 & 18.4 & 31.8 & 18.5 & 78.0 & 76.6 & 88.1 & 81.5 & 46.1 & \textbf{30.6} & \textbf{81.1} & \textbf{54.7} \\
\midrule
\multicolumn{14}{l}{\textit{$\sim$\,8B Models}} \\
\midrule
Ovis2-8B~\citep{lu2024ovis} & 56.4 & 70.9 & 26.3 & 40.9 & 27.6 & 84.7 & 83.3 & 94.2 & 89.1 & 56.4 & 41.4 & 87.8 & 63.7 \\
InternVL3-8B~\citep{zhu2025internvl3} & 57.8 & 70.6 & 26.0 & 45.9 & 31.3 & 86.1 & 82.2 & 92.0 & 88.1 & 57.8 & 43.4 & 87.1 & 64.4 \\
Qwen3-VL-8B-Instruct~\citep{bai2025qwen3} & 61.2 & 75.9 & 37.5 & 53.7 & 50.3 & 83.7 & 83.8 & 95.7 & 90.1 & 61.2 & 54.4 & 88.3 & 70.2 \\
InternVL3.5-8B-Pretrained~\citep{wang2025internvl3} & 56.1 & 62.0 & 24.0 & 40.0 & 33.1 & 84.4 & 78.5 & 91.7 & 80.8 & 56.1 & 39.8 & 83.9 & 61.2 \\
InternVL3.5-8B-Instruct~\citep{wang2025internvl3} & 58.1 & 70.8 & 35.5 & 48.1 & 32.5 & 86.3 & 77.2 & 91.7 & 82.9 & 58.1 & 46.7 & 84.5 & 64.8 \\
\midrule
\rowcolor{blue!8}Baseline ($P_v = 1, P_l = 1$) & 56.9 & 63.2 & 24.3 & 42.1 & 35.0 & 85.4 & 79.6 & 93.1 & 83.1 & 56.9 & 41.2 & 85.3 & 62.5 \\
\rowcolor{blue!8}ParVL-8B ($P_v = 2, P_l = 2$) & 57.8 & 64.4 & 26.6 & 41.6 & 35.2 & 84.7 & 80.3 & 93.2 & 83.4 & \textbf{57.8} & \textbf{42.0} & \textbf{85.4} & \textbf{63.0} \\
\bottomrule
\end{tabular}
}

\caption{\textbf{Comparison of multimodal general, mathematical, and OCR performance.} Within each model scale, the multi-branch ParVL variants are compared with a single-branch baseline ($P_v = 1, P_l = 1$) trained using the same SFT recipe; public models are shown for reference. \textbf{Bold} marks the best result among the displayed ParVL configurations of the same size. $P_v$ and $P_l$ denote the numbers of ViT and LLM branches, respectively.}
\label{tab:benchmark}
\end{table*}

\subsection{Setup}

\paragraph{Data.} Due to resource constraints, we use a 1/20 subsample of the InternVL3.5 SFT collection, comprising approximately 13B tokens and 4.6M instances~\citep{wang2025internvl3}. All ParVL variants are initialized from the corresponding InternVL3.5 pretrained checkpoint and undergo one epoch of full-parameter SFT. Within each model scale, all controlled branch comparisons use the same data, optimization recipe, and pretrained checkpoint. Full training and data-composition details are provided in the supplement.

\paragraph{Benchmarks.}
We evaluate nine benchmarks. MMMU$_{\text{dev-val}}$~\citep{Yue_2024_CVPR} assesses general multimodal reasoning. MathVista$_{\text{mini}}$~\citep{lu2024mathvista}, MathVision$_{\text{mini}}$~\citep{wang2024measuring}, WeMath~\citep{qiao2025we}, and LogicVista~\citep{xiao2024logicvista} evaluate mathematical and logical reasoning. ChartQA$_{\text{test}}$~\citep{masry2022chartqa}, TextVQA$_{\text{val}}$~\citep{singh2019textvqa}, DocVQA$_{\text{val}}$~\citep{mathew2021docvqa}, and OCRBench~\citep{liu2023ocrbench} assess chart, OCR, and document understanding. We report the MMMU score as General, average the four mathematical and logical reasoning benchmarks as Math, and average the four OCR/document benchmarks as OCR. ``All'' is the equal-weight average of all nine benchmark scores. We use the same evaluation protocol for all experiments.

\subsection{Main Results}

Table~\ref{tab:benchmark} compares ParVL with public MLLMs and same-recipe single-branch baselines. At 1B, the configuration ($P_v = 1, P_l = 4$) raises the overall average from 49.6 for the single-branch baseline ($P_v = 1, P_l = 1$) to 50.3, while jointly expanding both components to $4{:}4$ further raises it to 50.5. At 2B and 8B, the $2{:}2$ configuration raises the overall average from 54.4 to 54.7 and from 62.5 to 63.0, respectively. The same-recipe gains are therefore largest at 1B and smaller but positive at 2B and 8B. Public reference models provide additional context but use different training recipes.

Across scales, the most consistent domain-level gains occur in mathematical reasoning and OCR/document understanding, while General is more sensitive to branch allocation. As shown in Table~\ref{tab:vit_vs_llm}, the best 1B configurations raise Math from 26.0 to 28.0 at $4{:}4$ and OCR from 75.6 to 76.3 at $2{:}4$. General peaks at 41.1 under $2{:}2$ but falls below its 40.3 single-branch baseline under $1{:}4$ and $4{:}4$. At 2B, the evaluated $2{:}2$ configuration improves Math and OCR by 0.3 and 0.5, respectively, while General changes from 46.4 to 46.1. At 8B, the same configuration improves General, Math, and OCR by 0.9, 0.8, and 0.1, respectively. Prior controlled studies of looped language models find that parameter count tracks factual knowledge capacity, whereas additional computation through parameter reuse improves knowledge manipulation without increasing raw knowledge storage~\citep{zhu2025scaling}. ParVL similarly expands computation over shared backbone parameters, which may favor tasks requiring visual evidence integration and intermediate reasoning over knowledge-intensive benchmarks such as MMMU. The 8B result nevertheless shows that parallel scaling can also benefit broad-domain reasoning when the shared backbone has greater parameter capacity. Because all ParVL variants are trained on only 1/20 of the InternVL3.5 SFT collection, whether broader supervision strengthens the smaller-scale General results remains an open question.

\subsection{Ablation Results}

\begin{table}[t!]
\centering
\setlength{\tabcolsep}{2.0pt}
\renewcommand{\arraystretch}{0.82}
\scriptsize

\begin{subtable}[t]{0.32\linewidth}
    \centering
    \caption{MathVista$_{\text{mini}}$}
    \label{tab:ab_mathvista}
    \begin{tabular}{c c c c}
    \toprule
    & \multicolumn{3}{c}{$P_v$} \\
    \cmidrule(lr){2-4}
    $P_l$ & 1 & 2 & 4 \\
    \midrule
    $1$ & \cellcolor{blue!8!white}{43.9} & \cellcolor{blue!27!white}{45.8} & \cellcolor{blue!19!white}{45.0} \\
    $2$ & \cellcolor{blue!13!white}{44.4} & \cellcolor{blue!10!white}{44.1} & \cellcolor{blue!17!white}{44.8} \\
    $4$ & \cellcolor{blue!14!white}{44.5} & \cellcolor{blue!9!white}{44.0} & \cellcolor{blue!40!white}{48.0} \\
    \bottomrule
    \end{tabular}
\end{subtable}
\hfill
\begin{subtable}[t]{0.32\linewidth}
    \centering
    \caption{LogicVista}
    \label{tab:ab_logicvista}
    \begin{tabular}{c c c c}
    \toprule
    & \multicolumn{3}{c}{$P_v$} \\
    \cmidrule(lr){2-4}
    $P_l$ & 1 & 2 & 4 \\
    \midrule
    $1$ & \cellcolor{blue!20!white}{27.3} & \cellcolor{blue!9!white}{25.7} & \cellcolor{blue!40!white}{30.0} \\
    $2$ & \cellcolor{blue!27!white}{28.2} & \cellcolor{blue!16!white}{26.8} & \cellcolor{blue!21!white}{27.5} \\
    $4$ & \cellcolor{blue!35!white}{29.5} & \cellcolor{blue!25!white}{28.0} & \cellcolor{blue!33!white}{29.3} \\
    \bottomrule
    \end{tabular}
\end{subtable}
\hfill
\begin{subtable}[t]{0.32\linewidth}
    \centering
    \caption{WeMath}
    \label{tab:ab_wemath}
    \begin{tabular}{c c c c}
    \toprule
    & \multicolumn{3}{c}{$P_v$} \\
    \cmidrule(lr){2-4}
    $P_l$ & 1 & 2 & 4 \\
    \midrule
    $1$ & \cellcolor{blue!30!white}{11.6} & \cellcolor{blue!40!white}{12.3} & \cellcolor{blue!24!white}{11.3} \\
    $2$ & \cellcolor{blue!31!white}{11.7} & \cellcolor{blue!8!white}{10.1} & \cellcolor{blue!28!white}{11.5} \\
    $4$ & \cellcolor{blue!30!white}{11.6} & \cellcolor{blue!40!white}{12.3} & \cellcolor{blue!40!white}{12.3} \\
    \bottomrule
    \end{tabular}
\end{subtable}
\par\vspace{3.0mm}
\begin{subtable}[t]{0.32\linewidth}
    \centering
    \caption{ChartQA$_{\text{test}}$}
    \label{tab:ab_chartqa}
    \begin{tabular}{c c c c}
    \toprule
    & \multicolumn{3}{c}{$P_v$} \\
    \cmidrule(lr){2-4}
    $P_l$ & 1 & 2 & 4 \\
    \midrule
    $1$ & \cellcolor{blue!8!white}{71.7} & \cellcolor{blue!25!white}{72.9} & \cellcolor{blue!25!white}{72.9} \\
    $2$ & \cellcolor{blue!29!white}{73.2} & \cellcolor{blue!21!white}{72.6} & \cellcolor{blue!15!white}{72.2} \\
    $4$ & \cellcolor{blue!14!white}{72.1} & \cellcolor{blue!40!white}{74.0} & \cellcolor{blue!14!white}{72.1} \\
    \bottomrule
    \end{tabular}
\end{subtable}
\hfill
\begin{subtable}[t]{0.32\linewidth}
    \centering
    \caption{TextVQA$_{\text{val}}$}
    \label{tab:ab_textvqa}
    \begin{tabular}{c c c c}
    \toprule
    & \multicolumn{3}{c}{$P_v$} \\
    \cmidrule(lr){2-4}
    $P_l$ & 1 & 2 & 4 \\
    \midrule
    $1$ & \cellcolor{blue!20!white}{71.4} & \cellcolor{blue!20!white}{71.4} & \cellcolor{blue!24!white}{71.5} \\
    $2$ & \cellcolor{blue!16!white}{71.3} & \cellcolor{blue!20!white}{71.4} & \cellcolor{blue!8!white}{71.1} \\
    $4$ & \cellcolor{blue!28!white}{71.6} & \cellcolor{blue!36!white}{71.8} & \cellcolor{blue!40!white}{71.9} \\
    \bottomrule
    \end{tabular}
\end{subtable}
\hfill
\begin{subtable}[t]{0.32\linewidth}
    \centering
    \caption{OCRBench}
    \label{tab:ab_ocrbench}
    \begin{tabular}{c c c c}
    \toprule
    & \multicolumn{3}{c}{$P_v$} \\
    \cmidrule(lr){2-4}
    $P_l$ & 1 & 2 & 4 \\
    \midrule
    $1$ & \cellcolor{blue!33!white}{76.6} & \cellcolor{blue!26!white}{76.2} & \cellcolor{blue!17!white}{75.5} \\
    $2$ & \cellcolor{blue!12!white}{75.2} & \cellcolor{blue!22!white}{76.0} & \cellcolor{blue!22!white}{76.0} \\
    $4$ & \cellcolor{blue!31!white}{76.5} & \cellcolor{blue!26!white}{76.2} & \cellcolor{blue!40!white}{77.0} \\
    \bottomrule
    \end{tabular}
\end{subtable}
\par\vspace{3.0mm}
\begin{subtable}[t]{0.32\linewidth}
    \centering
    \caption{General}
    \label{tab:ab_general}
    \begin{tabular}{c c c c}
    \toprule
    & \multicolumn{3}{c}{$P_v$} \\
    \cmidrule(lr){2-4}
    $P_l$ & 1 & 2 & 4 \\
    \midrule
    $1$ & \cellcolor{blue!27!white}{40.3} & \cellcolor{blue!17!white}{39.9} & \cellcolor{blue!14!white}{39.7} \\
    $2$ & \cellcolor{blue!35!white}{40.8} & \cellcolor{blue!40!white}{41.1} & \cellcolor{blue!31!white}{40.6} \\
    $4$ & \cellcolor{blue!15!white}{39.8} & \cellcolor{blue!8!white}{39.1} & \cellcolor{blue!8!white}{39.0} \\
    \bottomrule
    \end{tabular}
\end{subtable}
\hfill
\begin{subtable}[t]{0.32\linewidth}
    \centering
    \caption{Avg. Math}
    \label{tab:ab_math}
    \begin{tabular}{c c c c}
    \toprule
    & \multicolumn{3}{c}{$P_v$} \\
    \cmidrule(lr){2-4}
    $P_l$ & 1 & 2 & 4 \\
    \midrule
    $1$ & \cellcolor{blue!16!white}{26.0} & \cellcolor{blue!21!white}{26.4} & \cellcolor{blue!30!white}{27.2} \\
    $2$ & \cellcolor{blue!23!white}{26.6} & \cellcolor{blue!9!white}{25.4} & \cellcolor{blue!8!white}{25.3} \\
    $4$ & \cellcolor{blue!35!white}{27.6} & \cellcolor{blue!19!white}{26.3} & \cellcolor{blue!40!white}{28.0} \\
    \bottomrule
    \end{tabular}
\end{subtable}
\hfill
\begin{subtable}[t]{0.32\linewidth}
    \centering
    \caption{Avg. OCR}
    \label{tab:ab_ocr}
    \begin{tabular}{c c c c}
    \toprule
    & \multicolumn{3}{c}{$P_v$} \\
    \cmidrule(lr){2-4}
    $P_l$ & 1 & 2 & 4 \\
    \midrule
    $1$ & \cellcolor{blue!12!white}{75.6} & \cellcolor{blue!16!white}{75.7} & \cellcolor{blue!8!white}{75.5} \\
    $2$ & \cellcolor{blue!8!white}{75.5} & \cellcolor{blue!12!white}{75.6} & \cellcolor{blue!8!white}{75.5} \\
    $4$ & \cellcolor{blue!16!white}{75.7} & \cellcolor{blue!40!white}{76.3} & \cellcolor{blue!24!white}{75.9} \\
    \bottomrule
\end{tabular}
\end{subtable}

\caption{\textbf{Ablation on branch-count configurations between vision branches ($P_v$) and language branches ($P_l$).} Darker cells indicate higher scores within each subtable.}
\label{tab:vit_vs_llm}
\end{table}

\begin{figure}[t]
\centering
\includegraphics[width=\linewidth]{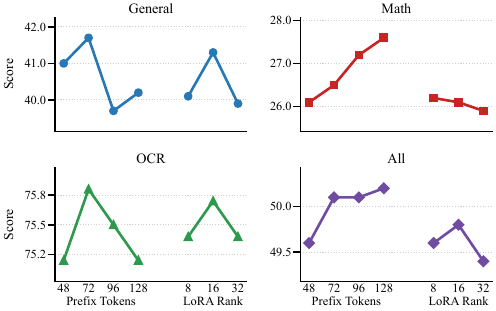}
\caption{\textbf{Comparison of two mechanisms for parameterizing parallel ViT branches} in ParVL-1B under $P_v = 4, P_l = 1$: branch-specific KV prefixes in prefix-conditioned attention and branch-specific LoRA modules. Prefix Tokens and LoRA Rank denote the KV-prefix length per ViT branch and the rank of each LoRA module, respectively. The language prefix remains fixed at 48 tokens~\citep{chen2025parallel}.}
\label{fig:vpt_lora_avg}
\end{figure}

\paragraph{Prefix vs. LoRA.}
Figure~\ref{fig:vpt_lora_avg} compares two implementations of vision-side parallelism. Prefix-conditioned attention differentiates parallel ViT branches using learned KV prefixes, whereas the LoRA variant assigns a branch-specific low-rank update to each parallel stream~\citep{hu2022lora}. Both implementations reuse the same ViT backbone under the same branch-count configuration ($P_v = 4, P_l = 1$) and a fixed 48-token language prefix. The comparison therefore isolates how new branch-specific parameters are introduced into the parallel vision streams. Across visual prefix lengths, 72 tokens perform best on General (41.7) and OCR (75.8), whereas 128 tokens achieve the highest Math (27.6) and overall (50.2) scores. We select 96 tokens as a reasoning-oriented, compute-aware compromise: it matches the 72-token overall score of 50.1 while improving Math by 0.7, and remains within 0.1 overall and 0.4 Math of the 128-token setting while using 25\% fewer visual prefix tokens.

Among the LoRA variants, rank 16 gives the highest General (41.3), OCR (75.7), and overall (49.8) scores. The best LoRA overall score remains 0.4 below the best prefix result, so we retain KV prefixes as ParVL's default parameterization.

\paragraph{Vision/language allocation.}
Table~\ref{tab:vit_vs_llm} reports nine ParVL-1B branch-count configurations. Relative to the single-branch configuration ($P_v = 1, P_l = 1$), the best allocations raise General from 40.3 to 41.1 at $2{:}2$, Math from 26.0 to 28.0 at $4{:}4$, and OCR from 75.6 to 76.3 at $2{:}4$. The largest $4{:}4$ configuration therefore does not perform best in every domain: it trails the best observed score by 2.1 on General and 0.4 on OCR. Thus, the observed allocation preferences cannot be explained by total branch count alone.

Allocation preferences also vary within the same domain. The complete heatmaps in the supplement show that MathVision favors the language-heavy $1{:}4$ configuration, LogicVista the vision-heavy $4{:}1$ configuration, and MathVista the jointly expanded $4{:}4$ configuration. These configurations improve upon $1{:}1$ by 3.6, 2.7, and 4.1, respectively. OCR/document benchmarks likewise split between $2{:}4$ for ChartQA and DocVQA and $4{:}4$ for TextVQA and OCRBench. These results indicate that domain labels do not fully explain the observed allocation preferences and instead suggest a task-level interaction between visual evidence processing and language-side reasoning. The nine-benchmark average likewise shows that increasing one branch axis does not guarantee improvement: at $P_l = 2$, increasing $P_v$ from 1 to 2 or 4 reduces the score from 49.9 to 49.4 and 49.3, whereas increasing $P_l$ to 4 recovers performance to 49.9 at $2{:}4$ and 50.5 at $4{:}4$. Benchmark workload descriptors are also provided in the supplement.

\subsection{Inference Efficiency}

\begin{figure}[t]
\centering
\includegraphics[width=\linewidth]{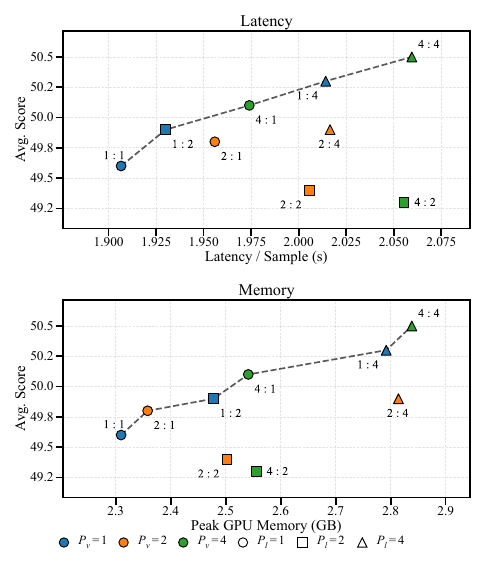}
\caption{\textbf{Performance--cost trade-offs} at batch size $=1$ for inputs containing 768 image-context tokens and 256 text tokens per sample (1024 input tokens total); generation is fixed to 128 tokens. Labels denote $P_v{:}P_l$, colors $P_v$, and marker shapes $P_l$. Dashed lines trace the performance--cost frontier in each panel. The reported $4{:}1$ configuration uses 96 visual prefix tokens per branch.}
\label{fig:efficiency_bs1}
\end{figure}

\begin{table}[t]
\centering
\setlength{\tabcolsep}{3.5pt}
\scriptsize
\begin{tabular}{@{}c c c c c c c@{}}
\toprule
\textbf{Trained $P_v{:}P_l$} & \textbf{Active $P_v{:}P_l$} & \textbf{Policy} & \textbf{General} & \textbf{Math} & \textbf{OCR} & \textbf{All} \\
\midrule
$4{:}4$ & $1{:}1$ & Random pair & 38.8 & 26.6 & 75.1 & 49.5 \\
$4{:}4$ & $1{:}1$ & Learned router & 38.8 & 27.5 & 75.8 & 50.2 \\
$4{:}4$ & $4{:}4$ & Aggregation & 39.0 & 28.0 & 75.9 & 50.5 \\
\bottomrule
\end{tabular}
\caption{\textbf{Exploration on sparse router module.} All rows use the same checkpoint trained under $P_v = 4, P_l = 4$. Random pair and the learned router each activate one ViT--LLM branch pair per sample, whereas Aggregation activates all branches.}
\label{tab:sparse_router}
\end{table}

\paragraph{Sparse routing.}
Inspired by the top-1 sparse-routing principle in MoE models~\citep{fedus2022switch}, Table~\ref{tab:sparse_router} evaluates whether a checkpoint trained under $P_v = 4, P_l = 4$ can be executed sparsely by activating only one ViT--LLM branch pair per sample. The learned router improves the overall average from 49.5 under random pair selection to 50.2, while dense $4{:}4$ aggregation reaches 50.5. Relative to random selection, the gains are concentrated in Math (26.6 to 27.5) and OCR (75.1 to 75.8), while General remains unchanged at 38.8. The learned router therefore operates within 0.3 of dense aggregation while reducing the active configuration from $4{:}4$ to $1{:}1$. The supplementary material details the router architecture, training objective, and inference procedure, and analyzes the empirical Math and OCR preferences of individual branches.

\begin{table}[t]
\centering
\setlength{\tabcolsep}{6.5pt}
\scriptsize
\begin{tabular}{@{}cclcccc@{}}
\toprule
\multirow{2}{*}{\textbf{$P_v{:}P_l$}} &
\multirow{2}{*}{\textbf{Shared KV}} &
\multirow{2}{*}{\textbf{Metric}} &
\multicolumn{4}{c}{\textbf{Batch Size}} \\
\cmidrule(lr){4-7}
& & & \textbf{1} & \textbf{2} & \textbf{4} & \textbf{8} \\
\midrule
\multirow{2}{*}{$4{:}4$} & \multirow{2}{*}{\ding{55}} &
Latency & $1.04{\times}$ & $1.09{\times}$ & $1.12{\times}$ & $1.22{\times}$ \\
& & Memory & $1.23{\times}$ & $1.41{\times}$ & $1.69{\times}$ & $2.12{\times}$ \\
\midrule
\multirow{2}{*}{$4{:}4$} & \multirow{2}{*}{\ding{51}} &
Latency & $1.23{\times}$ & $1.32{\times}$ & $1.37{\times}$ & $1.62{\times}$ \\
& & Memory & $1.11{\times}$ & $1.18{\times}$ & $1.30{\times}$ & $1.47{\times}$ \\
\bottomrule
\end{tabular}
\caption{\textbf{Relative inference costs.} Per-sample latency and peak
allocated GPU memory, normalized by standard $1{:}1$ at each batch size. Both
$4{:}4$ variants are profiled sequentially on the same NVIDIA H200 GPU.
Shared KV mean-reduces ordinary-token caches across LLM branches while
retaining branch-specific prefix K/V.}
\label{tab:efficiency_ratio}
\end{table}

\paragraph{Latency and memory.}
We profile the ParVL-1B variants with batch sizes $\{1,2,4,8\}$. Each sample contains 768 image-context tokens and 256 text tokens, with generation fixed to 128 tokens. After three untimed warm-up iterations to exclude one-time initialization overhead, we measure per-sample latency over ten timed iterations with CUDA synchronization and record peak allocated GPU memory; model loading and warm-up are excluded. The supplement details the input construction and memory protocol.

ParVL exposes parallelism within each request, providing additional hardware concurrency when opportunities for batching across requests are limited. Figure~\ref{fig:efficiency_bs1} presents the performance--cost trade-off at batch size 1. Dense $4{:}4$ aggregation improves the overall score from 49.6 to 50.5 while increasing per-sample latency and peak memory to $1.04{\times}$ and $1.23{\times}$ those of $1{:}1$, respectively. The $1{:}4$ configuration achieves 50.3 versus 50.5 for $4{:}4$, with lower latency and peak memory. As shown in Table~\ref{tab:efficiency_ratio}, the latency and memory ratios of standard $4{:}4$ rise to $1.22{\times}$ and $2.12{\times}$, respectively, as batch size increases from 1 to 8. The latency ratio grows more slowly than the memory ratio, consistent with concurrent branch execution partially overlapping the additional computation while branch-specific activations and separate KV caches for the LLM branches scale with batch size.

To reduce this memory growth, Shared KV retains branch-specific prefix K/V states while mean-reducing ordinary-token K/V states across the four LLM branches. As shown in Table~\ref{tab:efficiency_ratio}, at batch size 8, it lowers the memory ratio relative to standard $1{:}1$ from $2.12{\times}$ to $1.47{\times}$, but raises the latency ratio from $1.22{\times}$ to $1.62{\times}$. Shared KV incurs only a small decrease in the nine-benchmark average, with detailed per-benchmark accuracy results provided in the supplement. Overall, within-request parallelism is most attractive in low-concurrency settings, while Shared KV provides a memory--latency--accuracy trade-off as batching increases. The supplement provides implementation details, absolute measurements, and the full quality evaluation; all profiling results remain specific to this hardware and workload.

\section{Discussion and Limitations}
\label{sec:discussions}




\paragraph{Relationship to Ensemble Methods.}
From another perspective, ParVL can be viewed as an implicit ensemble of branch-wise next-token predictions~\citep{jacobs1991adaptive,shazeer2017outrageously,du2022glam}. Unlike conventional ensembles of independently parameterized models, ParVL uses lightweight, branch-specific conditioning to differentiate representations produced by shared backbones. This preserves the parameter efficiency of a shared model while producing branch-conditioned representations; the additional computation and memory are quantified in our efficiency analysis.

\paragraph{Comparison with MoE Models and Upcycling.}
MoE models expand capacity through independently parameterized experts and sparse token routing; upcycling initializes expert parameters from a dense checkpoint before allowing them to specialize during joint routed training~\citep{shazeer2017outrageously,fedus2022switch,li2024cumo}. ParVL instead keeps the ViT and LLM backbones shared and differentiates parallel streams through lightweight prefixes and aggregation modules, expanding computation with limited parameter growth. ParVL maintains a separate KV cache for each active LLM branch, whereas optional top-1 routing reduces execution to one ViT--LLM branch pair. Thus, MoE upcycling expands stored expert capacity, while ParVL expands and allocates shared-backbone computation.


\paragraph{Future Directions.}
ParVL currently expands parallel computation during SFT; extending parallel scaling to pretraining could reveal how joint vision--language scaling behavior emerges~\citep{chen2025parallel}. It is also complementary to token-efficient visual backbones based on adaptive token pruning or patch merging~\citep{ye2025atp,zhong2025aim,shao2025holitom,kim2024token}, which may offset the cost of multi-branch computation.

\paragraph{Limitations.}
We study only InternVL3.5 at 1B, 2B, and 8B and supervised fine-tuning rather than pretraining. Resource constraints limit training to a 1/20 subsample of the InternVL3.5 SFT collection with a single data mixture. Whether the modest gains at larger model scales partly reflect this restricted SFT budget remains to be tested through controlled data-scaling experiments. Validation across broader backbones, larger scales, and real deployment workloads remains future work.

\section{Conclusion}
\label{sec:conclusion}

In this paper, we introduce ParVL, a framework for parallel scaling that expands computation in MLLMs via parameter-shared, prefix-conditioned ViT and LLM branches. Controlled comparisons show that this parallel structure improves multimodal reasoning compared with same-recipe single-branch baselines while adding at most 4\% additional parameters across the evaluated configurations. We further conduct the first systematic study of computation allocation between the vision encoder ($P_v$) and language decoder ($P_l$). Our experiments reveal that no single branch-count configuration performs best across the evaluated tasks: reasoning performance benefits from expanding vision computation, language computation, or both, depending on task-specific multimodal requirements. These findings establish parallel scaling as a promising direction for improving MLLM reasoning capabilities and motivate task-aware, flexible allocation of computation between the two modalities.



\bibliography{aaai2027}

\appendix
\setcounter{secnumdepth}{2}
\setcounter{table}{0}
\setcounter{figure}{0}
\setcounter{equation}{0}
\counterwithin*{equation}{section}
\renewcommand{\thetable}{S\arabic{table}}
\renewcommand{\thefigure}{S\arabic{figure}}
\renewcommand{\theequation}{\thesection\arabic{equation}}

\par\medskip
\noindent
\begin{minipage}{\columnwidth}
\centering
\setlength{\tabcolsep}{3pt}
\begin{tabular}{@{}p{0.68\linewidth}r@{}}
\toprule
\textbf{Category} & \textbf{Approx. Share} \\
\midrule
Mathematical reasoning & 15\% \\
OCR and document understanding & 15\% \\
General multimodal and other data & 70\% \\
\bottomrule
\end{tabular}
\captionof{table}{\textbf{Approximate SFT data composition by instance count.}
Categories are estimated from task metadata and dataset names; percentages are
rounded.}
\label{tab:training_data_source}
\end{minipage}
\par\medskip

\section{Experimental Details}

\subsection{SFT Data and Recipe}
Table~\ref{tab:training_data_source} summarizes the approximate instance-level composition of the 4.62M-instance sampling manifest: 15\% mathematical reasoning, 15\% OCR and document understanding, and 70\% general multimodal and other instruction-following data. The categories are derived from task metadata and dataset names and are reported as rounded proportions.

\paragraph{Data availability.}
The 1/20 SFT subset is sampled from the InternVL3.5 SFT collection, which aggregates 1,301 data sources with heterogeneous access and redistribution conditions. Some component datasets are not publicly accessible or redistributable, so we report the subset size and aggregate domain composition rather than redistribute the underlying examples. Replacing these components with a fully public data mixture would change the supervision distribution and introduce a data-mixture confound into the controlled comparison of branch-count configurations. We consequently use the same sampled collection for all ParVL variants.


All ParVL models undergo one epoch of full-parameter SFT from the corresponding InternVL3.5 pretrained checkpoint. The global batch size is 512 and maximum sequence length is 32{,}768. We use a cosine schedule with 3\% warmup, peak learning rates of $8\times10^{-5}$ for branch-specific prefixes and aggregation modules and $8\times10^{-6}$ for pretrained parameters, weight decay 0.05, and \texttt{bf16}. Images use $448\times448$ tiles, dynamic tiling with up to 12 patches and thumbnail augmentation, and a 0.5 pixel-shuffle ratio. Training uses DeepSpeed ZeRO and gradient checkpointing.

\subsection{Compute Environment and Reproducibility}
Training and model-side evaluation run on nodes with eight NVIDIA H200 GPUs (143{,}771 MiB per GPU), two Intel Xeon Platinum 8558 CPUs, and 1.8 TiB of system memory. Each training worker reserves 40 CPU cores. Depending on model scale and branch configuration, distributed SFT uses 4--16 such nodes (32--128 GPUs); the fixed-workload latency and memory profile uses one H200 from the same hardware pool. The software environment is Ubuntu 24.04.1 LTS with Python 3.12.3, PyTorch 2.6.0+\texttt{cu126}, CUDA Toolkit 12.8, Transformers 4.55.0, DeepSpeed 0.17.4, FlashAttention 3.0.0b1, Accelerate 1.7.0, and Datasets 3.6.0. All SFT and router-training jobs use random seed 42. Benchmark generation is deterministic (\texttt{do\_sample=False}).

\begin{table*}[t]
\centering
\small
\setlength{\tabcolsep}{6pt}
\begin{tabular*}{\textwidth}{@{\extracolsep{\fill}} l c c c c}
\toprule
Dataset &
\makecell{Avg. Source\\Pixels} &
\makecell{Avg. Question\\Tokens} &
\makecell{Avg. Reference\\Tokens} &
\makecell{Best Empirical\\$P_v{:}P_l$} \\
\midrule
MMMU$_{\text{dev-val}}$ & 470,496 & 50.6 & 22.9 & $2{:}2$ \\
MathVista$_{\text{mini}}$ & 431,255 & 67.3 & 2.2 & $4{:}4$ \\
MathVision$_{\text{mini}}$ & 479,278 & 97.4 & 1.2 & $1{:}4$ \\
LogicVista & 383,914 & 53.6 & 65.4 & $4{:}1$ \\
WeMath & 231,998 & 36.4 & 1.0 & $2{:}1$ \\
ChartQA$_{\text{test}}$ & 454,992 & 15.5 & 3.5 & $2{:}4$ \\
TextVQA$_{\text{val}}$ & 770,512 & 8.3 & 49.9 & $4{:}4$ \\
DocVQA$_{\text{val}}$ & 3,925,844 & 11.0 & 11.0 & $2{:}4$ \\
OCRBench & 1,017,574 & 12.5 & 11.0 & $4{:}4$ \\
\bottomrule
\end{tabular*}
\caption{\textbf{Dataset workload descriptors and empirically best branch-count configurations.} Best configurations are selected from Table~\ref{tab:appendix_full_compute_allocation}; ties favor the lower-FLOP configuration.}
\label{tab:pixels_vs_tokens}
\end{table*}

\begin{table*}[t]
\centering
\setlength{\tabcolsep}{7.5pt}
\begin{tabular}{l c cccccc}
\toprule
\multirow{2}{*}{\textbf{Backbone}} & \multirow{2}{*}{\textbf{Config.} ($P_v{:}P_l$)} & \multicolumn{2}{c}{\textbf{ViT Part}} & \multicolumn{2}{c}{\textbf{LLM Part}} & \multicolumn{2}{c}{\textbf{Total}} \\
\cmidrule(lr){3-4} \cmidrule(lr){5-6} \cmidrule(lr){7-8}
& & FLOPs (G) & Params (M) & FLOPs (G) & Params (M) & FLOPs (G) & Params (M) \\
\midrule
\multirow{9}{*}{1B}
& $1{:}1$ & 1861.71 & 304.01 & 1373.25 & 751.63 & 3243.03 & 1060.90 \\
& $1{:}2$ & 1861.71 & 304.01 & 2392.87 & 759.24 & 4262.65 & 1068.50 \\
& $1{:}4$ & 1861.71 & 304.01 & 4427.28 & 766.84 & 6297.06 & 1076.11 \\
\cmidrule(l){2-8}
& $2{:}1$ & 3732.62 & 315.55 & 1373.25 & 751.63 & 5113.94 & 1072.44 \\
& $2{:}2$ & 3732.62 & 315.55 & 2392.87 & 759.24 & 6133.56 & 1080.04 \\
& $2{:}4$ & 3732.62 & 315.55 & 4427.28 & 766.84 & 8167.96 & 1087.64 \\
\cmidrule(l){2-8}
& $4{:}1$ & 7461.52 & 327.09 & 1373.25 & 751.63 & 8842.83 & 1083.97 \\
& $4{:}2$ & 7461.52 & 327.09 & 2392.87 & 759.24 & 9862.46 & 1091.58 \\
& $4{:}4$ & 7461.52 & 327.09 & 4427.28 & 766.84 & 11896.86 & 1099.18 \\
\midrule
\multirow{2}{*}{2B}
& $1{:}1$ & 1861.71 & 304.01 & 3964.12 & 2031.74 & 5845.17 & 2348.35 \\
& $2{:}2$ & 3732.62 & 315.55 & 7230.65 & 2045.64 & 10982.61 & 2373.78 \\
\midrule
\multirow{2}{*}{8B}
& $1{:}1$ & 1861.71 & 304.01 & 17437.41 & 8190.74 & 19350.68 & 8528.32 \\
& $2{:}2$ & 3732.62 & 315.55 & 33518.30 & 8231.38 & 37302.47 & 8580.50 \\
\bottomrule
\end{tabular}
\caption{\textbf{Computational Cost and Parameter Analysis.} We report the InternVL3.5-1B, InternVL3.5-2B, and InternVL3.5-8B ParVL checkpoints used in our experiments. FLOPs are fixed-length forward-pass estimates at batch size 1 for 768 image-context tokens, 256 text tokens, and 128 response tokens. The projector cost is included in the \textit{Total}: 8.07 GFLOPs / 5.25 M parameters for 1B, 19.34 GFLOPs / 12.60 M parameters for 2B, and 51.56 GFLOPs / 33.57 M parameters for 8B.}
\label{tab:total_param_and_flops}
\vspace{-16pt}
\end{table*}

\subsection{Evaluation Metrics and Score Aggregation}
We use the benchmark-native evaluators in VLMEvalKit~\citep{duan2024vlmevalkit} and place every reported score on a 0--100 scale. MMMU$_{\text{dev-val}}$~\citep{Yue_2024_CVPR} reports example-level accuracy on its validation split. MathVista$_{\text{mini}}$~\citep{lu2024mathvista} reports accuracy after extracting the final answer and normalizing it according to the required answer type (multiple choice, integer, or floating point). MathVision$_{\text{mini}}$~\citep{wang2024measuring} likewise reports accuracy after answer extraction. Multiple-choice predictions must match the correct option, while free-form answers are evaluated by exact match or, for numerical answers, equivalence within a tolerance of $10^{-6}$. LogicVista~\citep{xiao2024logicvista} uses exact set match between the extracted and reference option letters, so a multi-answer prediction is correct only when it contains all and only the reference choices.

For WeMath~\citep{qiao2025we}, we report its official strict score over the 525 underlying problem groups. Let $N_{\mathrm{CM}}$ be the number of groups for which every constituent one-step question and the combined multi-step question are correct (complete mastery), and let $N_{\mathrm{IG}}$ be the number for which all constituent questions are correct but the combined question is wrong (inadequate generalization). The reported score is
\begin{equation}
S_{\mathrm{WeMath}}^{\mathrm{strict}}
=100\frac{N_{\mathrm{CM}}+0.5N_{\mathrm{IG}}}{525}.
\end{equation}
Groups with any incorrect constituent answer receive no strict-score credit, including cases in which the combined answer happens to be correct.

TextVQA~\citep{singh2019textvqa} uses the standard VQA human-consensus score. For normalized prediction $\hat a_i$ and $m_i$ normalized references $a_{ij}$, its per-example score is
\begin{equation}
v_i=\frac{1}{m_i}\sum_{j=1}^{m_i}
\min\left(1,\frac{1}{3}\sum_{k\ne j}
\mathds{1}[\hat a_i=a_{ik}]\right).
\end{equation}
ChartQA~\citep{masry2022chartqa} uses relaxed accuracy: nonnumeric answers require case-insensitive exact match, whereas numeric answers receive credit under the evaluator's 5\% relative-tolerance rule. DocVQA~\citep{mathew2021docvqa} uses average normalized Levenshtein similarity (ANLS). For the best-matching reference, similarity is one minus normalized edit distance; values below 0.5 are set to zero. OCRBench~\citep{liu2023ocrbench} counts an example as correct when any normalized reference occurs in the prediction (with whitespace removed for handwritten mathematical expressions), sums these hits over its 1{,}000 examples, and divides the total by 10 to obtain a 0--100 score.

MMMU, MathVista, MathVision, and WeMath first attempt deterministic answer extraction. Unresolved cases, and all LogicVista cases, use \texttt{gpt-4o-2024-11-20} as a fixed answer extractor. It maps a generated response to an option or short answer; it does not replace the task-specific scoring rules above. ChartQA, TextVQA, DocVQA, and OCRBench use only deterministic evaluators.

Let $s_b$ denote the 0--100 score for benchmark $b$, let
$\mathcal{B}_{\mathrm{Math}}=\{\text{MathVista},\text{MathVision},\text{WeMath},\text{LogicVista}\}$, and let
$\mathcal{B}_{\mathrm{OCR}}=\{\text{ChartQA},\text{TextVQA},\text{DocVQA},\text{OCRBench}\}$. The aggregate scores are
\begin{equation}
\begin{aligned}
S_{\mathrm{General}} &= s_{\mathrm{MMMU}},\\
S_{\mathrm{Math}} &= \frac{1}{4}\sum_{b\in\mathcal{B}_{\mathrm{Math}}}s_b,\qquad
S_{\mathrm{OCR}} = \frac{1}{4}\sum_{b\in\mathcal{B}_{\mathrm{OCR}}}s_b,\\
S_{\mathrm{All}} &= \frac{1}{9}\left(
s_{\mathrm{MMMU}}+
\sum_{b\in\mathcal{B}_{\mathrm{Math}}\cup\mathcal{B}_{\mathrm{OCR}}}s_b
\right).
\end{aligned}
\end{equation}
We retain each benchmark's native metric to remain comparable with prior work and to respect its answer format, including annotator consensus, numerical tolerance, edit similarity, and structured multiple-choice reasoning. We use equal benchmark weights rather than pooling examples so that a larger dataset cannot dominate an aggregate. The General, Math, and OCR groupings expose the domain-level allocation effects central to our study, while the nine-task average provides one compact overall summary.

\subsection{Number of Runs}
Each reported model-quality result comes from one training run and one benchmark evaluation per configuration; model generation uses deterministic decoding, and we do not average results across random seeds. The sparse-router result likewise comes from one router-training run. Parameter counts and FLOP estimates are deterministic calculations performed once per configuration. For inference efficiency, each reported latency and peak-memory value is computed from ten timed iterations after three untimed warm-up iterations, as detailed in the profiling protocol below.

\section{Dataset and Compute Analyses}

\subsection{Dataset Workload Descriptors}
Table~\ref{tab:pixels_vs_tokens} reports average source-image pixels, question and reference lengths, and the empirically best branch allocation. These pre-tiling and pre-prompt descriptors are not FLOP measurements or causal explanations of allocation. Ties in the final column use the lowest-FLOP choice. Table~\ref{tab:appendix_full_compute_allocation} provides the full heatmaps.

Table~\ref{tab:pixels_vs_tokens} reveals that allocation preferences are frequently asymmetric. MathVision, LogicVista, ChartQA, and DocVQA each have a uniquely best asymmetric configuration, while WeMath selects an asymmetric configuration under the stated lowest-FLOP tie-breaking rule. Moreover, the empirically best configurations for all four OCR/document benchmarks use $P_l=4$, although their preferred vision branch counts differ. No clear correspondence is evident between the coarse workload descriptors and the selected allocations, suggesting that image size and text length alone do not explain the observed preferences.

Question lengths tokenize \texttt{question}; reference lengths tokenize \texttt{explanation} for MMMU, \texttt{reasoning} for LogicVista, and \texttt{answer} otherwise with the Qwen3-VL-2B-Instruct \texttt{Qwen2Tokenizer}~\citep{bai2025qwen3} (\texttt{use\_fast=False}, no special tokens). Empty references are encoded as empty strings and contribute zero tokens; this affects 831/1{,}050 MMMU examples.

\subsection{Parameter and FLOP Analysis}
Table~\ref{tab:total_param_and_flops} reports parameter and FLOP estimates under the same fixed workload as the efficiency profile; totals include the shared projector. Parameters are counted directly from checkpoint tensor shapes. Following our \texttt{calflops} convention, we sum separate ViT, projector, and LLM forward-pass estimates, with the LLM evaluated at a sequence length of 1{,}152 tokens, matching the token-length profile used for inference profiling. Branches reuse backbone parameters, but computation still increases with $P_v$ and $P_l$.

Relative to $1{:}1$ at each scale, the largest evaluated configuration
increases total parameters by 3.6\% at 1B ($4{:}4$), 1.1\% at 2B
($2{:}2$), and 0.6\% at 8B ($2{:}2$), while increasing total FLOPs by
266.8\%, 87.9\%, and 92.8\%, respectively.

\subsection{Full Compute Allocation}

Table~\ref{tab:appendix_full_compute_allocation} reports the complete
benchmark-level score grids for all evaluated branch-count
configurations.

\FloatBarrier

\subsection{Decoded Output Length}
To examine whether ParVL changes the length of generated responses, we
retokenize the saved prediction text with the shared Qwen2 tokenizer using
\texttt{add\_special\_tokens=False}, thereby excluding tokenizer-added BOS/EOS
tokens. Any textual tags present in the saved predictions would still be
counted. No explicit \texttt{<think>} tags appear in the saved predictions, so
we measure the length of the entire decoded response rather than reasoning-token
length. Under the 4{,}096-token generation limit, responses with at least
4{,}090 retokenized tokens are treated as near-limit cases and excluded from
the means.

\begin{table}[!t]
\centering
\setlength{\tabcolsep}{2.0pt}
\scriptsize

\begin{subtable}[t]{0.32\linewidth}
    \centering
    \caption{MMMU$_{\text{dev-val}}$}
    \label{tab:app_mmmu}
    \begin{tabular}{c c c c}
    \toprule
    & \multicolumn{3}{c}{$P_v$} \\
    \cmidrule(lr){2-4}
    $P_l$ & 1 & 2 & 4 \\
    \midrule
    $1$ & \cellcolor{blue!27!white}{40.3} & \cellcolor{blue!17!white}{39.9} & \cellcolor{blue!14!white}{39.7} \\
    $2$ & \cellcolor{blue!35!white}{40.8} & \cellcolor{blue!40!white}{41.1} & \cellcolor{blue!31!white}{40.6} \\
    $4$ & \cellcolor{blue!15!white}{39.8} & \cellcolor{blue!8!white}{39.1} & \cellcolor{blue!8!white}{39.0} \\
    \bottomrule
    \end{tabular}
\end{subtable}
\hfill
\begin{subtable}[t]{0.32\linewidth}
    \centering
    \caption{MathVista$_{\text{mini}}$}
    \label{tab:app_mathvista}
    \begin{tabular}{c c c c}
    \toprule
    & \multicolumn{3}{c}{$P_v$} \\
    \cmidrule(lr){2-4}
    $P_l$ & 1 & 2 & 4 \\
    \midrule
    $1$ & \cellcolor{blue!8!white}{43.9} & \cellcolor{blue!27!white}{45.8} & \cellcolor{blue!19!white}{45.0} \\
    $2$ & \cellcolor{blue!13!white}{44.4} & \cellcolor{blue!10!white}{44.1} & \cellcolor{blue!17!white}{44.8} \\
    $4$ & \cellcolor{blue!14!white}{44.5} & \cellcolor{blue!9!white}{44.0} & \cellcolor{blue!40!white}{48.0} \\
    \bottomrule
    \end{tabular}
\end{subtable}
\hfill
\begin{subtable}[t]{0.32\linewidth}
    \centering
    \caption{MathVision$_{\text{mini}}$}
    \label{tab:app_mathvision}
    \begin{tabular}{c c c c}
    \toprule
    & \multicolumn{3}{c}{$P_v$} \\
    \cmidrule(lr){2-4}
    $P_l$ & 1 & 2 & 4 \\
    \midrule
    $1$ & \cellcolor{blue!24!white}{21.1} & \cellcolor{blue!27!white}{21.7} & \cellcolor{blue!30!white}{22.4} \\
    $2$ & \cellcolor{blue!28!white}{22.0} & \cellcolor{blue!21!white}{20.4} & \cellcolor{blue!8!white}{17.4} \\
    $4$ & \cellcolor{blue!40!white}{24.7} & \cellcolor{blue!22!white}{20.7} & \cellcolor{blue!30!white}{22.4} \\
    \bottomrule
    \end{tabular}
\end{subtable}
\par\vspace{2mm}

\begin{subtable}[t]{0.32\linewidth}
    \centering
    \caption{LogicVista}
    \label{tab:app_logicvista}
    \begin{tabular}{c c c c}
    \toprule
    & \multicolumn{3}{c}{$P_v$} \\
    \cmidrule(lr){2-4}
    $P_l$ & 1 & 2 & 4 \\
    \midrule
    $1$ & \cellcolor{blue!20!white}{27.3} & \cellcolor{blue!9!white}{25.7} & \cellcolor{blue!40!white}{30.0} \\
    $2$ & \cellcolor{blue!27!white}{28.2} & \cellcolor{blue!16!white}{26.8} & \cellcolor{blue!21!white}{27.5} \\
    $4$ & \cellcolor{blue!35!white}{29.5} & \cellcolor{blue!25!white}{28.0} & \cellcolor{blue!33!white}{29.3} \\
    \bottomrule
    \end{tabular}
\end{subtable}
\hfill
\begin{subtable}[t]{0.32\linewidth}
    \centering
    \caption{WeMath}
    \label{tab:app_wemath}
    \begin{tabular}{c c c c}
    \toprule
    & \multicolumn{3}{c}{$P_v$} \\
    \cmidrule(lr){2-4}
    $P_l$ & 1 & 2 & 4 \\
    \midrule
    $1$ & \cellcolor{blue!30!white}{11.6} & \cellcolor{blue!40!white}{12.3} & \cellcolor{blue!24!white}{11.3} \\
    $2$ & \cellcolor{blue!31!white}{11.7} & \cellcolor{blue!8!white}{10.1} & \cellcolor{blue!28!white}{11.5} \\
    $4$ & \cellcolor{blue!30!white}{11.6} & \cellcolor{blue!40!white}{12.3} & \cellcolor{blue!40!white}{12.3} \\
    \bottomrule
    \end{tabular}
\end{subtable}
\hfill
\begin{subtable}[t]{0.32\linewidth}
    \centering
    \caption{ChartQA$_{\text{test}}$}
    \label{tab:app_chartqa}
    \begin{tabular}{c c c c}
    \toprule
    & \multicolumn{3}{c}{$P_v$} \\
    \cmidrule(lr){2-4}
    $P_l$ & 1 & 2 & 4 \\
    \midrule
    $1$ & \cellcolor{blue!8!white}{71.7} & \cellcolor{blue!25!white}{72.9} & \cellcolor{blue!25!white}{72.9} \\
    $2$ & \cellcolor{blue!29!white}{73.2} & \cellcolor{blue!21!white}{72.6} & \cellcolor{blue!15!white}{72.2} \\
    $4$ & \cellcolor{blue!14!white}{72.1} & \cellcolor{blue!40!white}{74.0} & \cellcolor{blue!14!white}{72.1} \\
    \bottomrule
    \end{tabular}
\end{subtable}
\par\vspace{2mm}

\begin{subtable}[t]{0.32\linewidth}
    \centering
    \caption{TextVQA$_{\text{val}}$}
    \label{tab:app_textvqa}
    \begin{tabular}{c c c c}
    \toprule
    & \multicolumn{3}{c}{$P_v$} \\
    \cmidrule(lr){2-4}
    $P_l$ & 1 & 2 & 4 \\
    \midrule
    $1$ & \cellcolor{blue!20!white}{71.4} & \cellcolor{blue!20!white}{71.4} & \cellcolor{blue!24!white}{71.5} \\
    $2$ & \cellcolor{blue!16!white}{71.3} & \cellcolor{blue!20!white}{71.4} & \cellcolor{blue!8!white}{71.1} \\
    $4$ & \cellcolor{blue!28!white}{71.6} & \cellcolor{blue!36!white}{71.8} & \cellcolor{blue!40!white}{71.9} \\
    \bottomrule
    \end{tabular}
\end{subtable}
\hfill
\begin{subtable}[t]{0.32\linewidth}
    \centering
    \caption{DocVQA$_{\text{val}}$}
    \label{tab:app_docvqa}
    \begin{tabular}{c c c c}
    \toprule
    & \multicolumn{3}{c}{$P_v$} \\
    \cmidrule(lr){2-4}
    $P_l$ & 1 & 2 & 4 \\
    \midrule
    $1$ & \cellcolor{blue!21!white}{82.6} & \cellcolor{blue!14!white}{82.4} & \cellcolor{blue!8!white}{82.2} \\
    $2$ & \cellcolor{blue!14!white}{82.4} & \cellcolor{blue!18!white}{82.5} & \cellcolor{blue!27!white}{82.8} \\
    $4$ & \cellcolor{blue!14!white}{82.4} & \cellcolor{blue!40!white}{83.2} & \cellcolor{blue!24!white}{82.7} \\
    \bottomrule
    \end{tabular}
\end{subtable}
\hfill
\begin{subtable}[t]{0.32\linewidth}
    \centering
    \caption{OCRBench}
    \label{tab:app_ocrbench}
    \begin{tabular}{c c c c}
    \toprule
    & \multicolumn{3}{c}{$P_v$} \\
    \cmidrule(lr){2-4}
    $P_l$ & 1 & 2 & 4 \\
    \midrule
    $1$ & \cellcolor{blue!33!white}{76.6} & \cellcolor{blue!26!white}{76.2} & \cellcolor{blue!17!white}{75.5} \\
    $2$ & \cellcolor{blue!12!white}{75.2} & \cellcolor{blue!22!white}{76.0} & \cellcolor{blue!22!white}{76.0} \\
    $4$ & \cellcolor{blue!31!white}{76.5} & \cellcolor{blue!26!white}{76.2} & \cellcolor{blue!40!white}{77.0} \\
    \bottomrule
\end{tabular}
\end{subtable}

\caption{\textbf{Full benchmark-level compute allocation heatmaps.} Rows denote language branch counts ($P_l$), columns denote vision branch counts ($P_v$), and darker cells indicate higher scores within each benchmark.}
\label{tab:appendix_full_compute_allocation}
\end{table}

\begin{table*}[t]
\centering
\fontsize{8}{10}\selectfont
\setlength{\tabcolsep}{1.1mm}
\renewcommand{\arraystretch}{1.12}
\resizebox{\linewidth}{!}{
\begin{tabular}{l|ccccccccc|cccc}
\toprule
\multirow{2}{*}{\textbf{Statistic}}
 & \multicolumn{1}{c}{\textbf{General}} & \multicolumn{4}{c}{\textbf{Math}}
 & \multicolumn{4}{c|}{\textbf{OCR}} & \multicolumn{4}{c}{\textbf{Average}} \\
\cmidrule(lr){2-2} \cmidrule(lr){3-6} \cmidrule(lr){7-10} \cmidrule(lr){11-14}
 & MMMU$_{\text{dev-val}}$ & MathVista$_{\text{mini}}$ & MathVision$_{\text{mini}}$ & LogicVista & WeMath
 & ChartQA$_{\text{test}}$ & TextVQA$_{\text{val}}$ & DocVQA$_{\text{val}}$ & OCRBench
 & General & Math & OCR & \textbf{All} \\
\midrule
Clean mean ($1{:}1$)
& 1.2 & 38.9 & 60.3 & 62.0 & 1.0
& 3.4 & 3.5 & 5.1 & 5.7
& 1.2 & 40.6 & 4.4 & 20.1 \\
Clean mean ($4{:}4$)
& 1.2 & 41.5 & 78.8 & 66.6 & 1.0
& 3.4 & 2.8 & 5.1 & 5.9
& 1.2 & 47.0 & 4.3 & 22.9 \\
Clean paired $\Delta$ ($4{:}4-1{:}1$)
& $+0.0$ & $+2.7$ & $+18.1$ & $+5.1$ & $+0.0$
& $+0.0$ & $-0.6$ & $+0.0$ & $+0.2$
& $+0.0$ & $+6.5$ & $-0.1$ & $+2.8$ \\
\bottomrule
\end{tabular}
}
\caption{\textbf{Decoded output length for ParVL-1B.} Mean lengths exclude
outputs at or above the near-cap threshold. Paired differences retain only
examples below the threshold for both configurations. Domain and overall columns
average the benchmark-level means with equal weight.}
\label{tab:decoded_output_length}
\end{table*}

\begin{table*}[t]
\centering
\fontsize{8}{10}\selectfont
\setlength{\tabcolsep}{1.1mm}
\renewcommand{\arraystretch}{1.12}
\resizebox{\linewidth}{!}{
\begin{tabular}{l|ccccccccc|cccc}
\toprule
\multirow{2}{*}{\textbf{KV Cache}}
 & \multicolumn{1}{c}{\textbf{General}} & \multicolumn{4}{c}{\textbf{Math}}
 & \multicolumn{4}{c|}{\textbf{OCR}} & \multicolumn{4}{c}{\textbf{Average}} \\
\cmidrule(lr){2-2} \cmidrule(lr){3-6} \cmidrule(lr){7-10} \cmidrule(lr){11-14}
 & MMMU$_{\text{dev-val}}$ & MathVista$_{\text{mini}}$ & MathVision$_{\text{mini}}$ & LogicVista & WeMath
 & ChartQA$_{\text{test}}$ & TextVQA$_{\text{val}}$ & DocVQA$_{\text{val}}$ & OCRBench
 & General & Math & OCR & \textbf{All} \\
\midrule
Standard & 39.0 & 48.0 & 22.4 & 29.3 & 12.3 & 72.1 & 71.9 & 82.7 & 77.0
& 39.0 & 28.0 & 75.9 & 50.5 \\
Shared & 39.8 & 47.1 & 19.4 & 30.6 & 11.8 & 72.0 & 71.7 & 82.8 & 77.1
& 39.8 & 27.2 & 75.9 & 50.3 \\
\midrule
$\Delta$ (Shared $-$ Standard)
& $+0.8$ & $-0.9$ & $-3.0$ & $+1.3$ & $-0.5$
& $-0.1$ & $-0.2$ & $+0.1$ & $+0.1$
& $+0.8$ & $-0.8$ & $0.0$ & $-0.2$ \\
\bottomrule
\end{tabular}
}
\caption{\textbf{Per-benchmark accuracy with Shared KV.} Both rows use the
same checkpoint trained with $P_v = 4, P_l = 4$; only the inference-time KV
cache changes. Both variants use the same
\texttt{gpt-4o-2024-11-20} answer extractor wherever API extraction is
required. All is the equal-weight average over the nine displayed benchmarks.}
\label{tab:shared_kv_accuracy}
\end{table*}

Table~\ref{tab:decoded_output_length} compares the single-branch baseline with
dense $4{:}4$ ParVL. Averaging the mean response length equally across
MathVista, MathVision, LogicVista, and WeMath, the value increases from 40.6 to
47.0 tokens, with an average paired difference of 6.5 tokens. This difference
is concentrated on MathVision, whereas General and OCR remain nearly unchanged. A
10{,}000-resample benchmark-balanced paired bootstrap (seed 42) gives a 95\%
interval of approximately $[-3.1,13.3]$ tokens for the paired difference.
Because this interval includes zero, the analysis does not establish
a statistically reliable overall increase in response length. The observed
accuracy gains therefore do not coincide with a uniform increase in decoded
output length.

\section{Inference Efficiency}

\subsection{Inference Profiling Protocol}
We profile the ParVL-1B variants at training step 7{,}000 using the batched generation path. All latency and memory measurements are collected on a single NVIDIA H200 GPU using \texttt{bf16} inference. Each sample contains 768 image-context tokens and 256 text tokens, with generation fixed to 128 tokens. We create one synthetic $448{\times}448$ RGB patch and repeat it three times; each copy produces 256 image-context tokens, yielding 768 in total. The text prompt is extended with fixed filler words until its tokenizer-observed non-image input length reaches 256 tokens. Generation sets \texttt{do\_sample=False} and \texttt{min\_new\_tokens=max\_new\_tokens=128}.

For every profiled branch-count configuration and each batch size in $\{1,2,4,8\}$, we run three untimed warm-up iterations followed by ten timed iterations, with CUDA synchronization around timing. Per-sample latency divides elapsed time by the iteration count and batch size. Peak-memory statistics are reset immediately before timed iterations; we report peak allocated GPU memory, including resident model allocations and runtime activations/KV caches.

\subsection{Shared-KV Cache Analysis}
The standard $4{:}4$ inference path persistently stores ordinary-token
key and value states for all four LLM branches. Shared KV retains the learned
prefix K/V states separately for each branch but shares the ordinary-token
cache. During prompt prefill, the ordinary-token K/V states at each decoder
layer are mean-reduced across branches in \texttt{fp32}, cast back to the cache
dtype, and stored once per sample. During autoregressive decoding, the K/V
states of each new decoding token are reduced in the same way and appended to
the shared per-layer cache. For attention, the shared ordinary-token cache is
temporarily expanded across branches and concatenated with the corresponding
branch-specific prefix K/V states. Because the branch hidden states diverge
after the first decoder layer, this procedure is an inference-time
approximation rather than an algebraically equivalent cache reorganization.

Table~\ref{tab:shared_kv_accuracy} shows that the approximation has mixed
task-level effects. MMMU and LogicVista increase by 0.8 and 1.3 points,
respectively, whereas MathVision decreases by 3.0 points. The OCR benchmarks
change by at most 0.2 points, and the equal-weight nine-benchmark average
decreases by 0.2 points.

\begin{table}[t]
\centering
\setlength{\tabcolsep}{6.5pt}
\scriptsize
\begin{tabular}{@{}cclcccc@{}}
\toprule
\multirow{2}{*}{\textbf{$P_v{:}P_l$}} &
\multirow{2}{*}{\textbf{Shared KV}} &
\multirow{2}{*}{\textbf{Metric}} &
\multicolumn{4}{c}{\textbf{Batch Size}} \\
\cmidrule(lr){4-7}
& & & \textbf{1} & \textbf{2} & \textbf{4} & \textbf{8} \\
\midrule
\multirow{2}{*}{$1{:}1$} & \multirow{2}{*}{\ding{55}} &
Latency (s) & 1.963 & 0.977 & 0.498 & 0.256 \\
& & Memory (GB) & 2.310 & 2.458 & 2.760 & 3.361 \\
\midrule
\multirow{2}{*}{$4{:}4$} & \multirow{2}{*}{\ding{55}} &
Latency (s) & 2.048 & 1.061 & 0.558 & 0.313 \\
& & Memory (GB) & 2.837 & 3.456 & 4.677 & 7.117 \\
\midrule
\multirow{2}{*}{$4{:}4$} & \multirow{2}{*}{\ding{51}} &
Latency (s) & 2.422 & 1.288 & 0.681 & 0.414 \\
& & Memory (GB) & 2.571 & 2.908 & 3.585 & 4.927 \\
\bottomrule
\end{tabular}
\caption{\textbf{Absolute Shared-KV profiling results.} Latency is measured in
seconds per sample, and memory is peak allocated GPU memory in GB. The three
inference settings are profiled sequentially on the same NVIDIA H200 GPU.}
\label{tab:shared_kv_absolute_efficiency}
\end{table}

Table~\ref{tab:shared_kv_absolute_efficiency} gives the absolute measurements
underlying the relative inference-cost comparison in the main paper. Relative to standard $4{:}4$, Shared KV
reduces total peak allocated memory by 9.4\%, 15.8\%, 23.3\%, and 30.8\% at
batch sizes 1, 2, 4, and 8, respectively, while increasing per-sample latency
by 18.3\%, 21.4\%, 22.1\%, and 32.3\%. After subtracting the resident
allocation measured immediately before timed generation, the incremental peak
memory reduction is 44.5--45.1\% across batch sizes. This is smaller than the
75\% reduction in the persistent ordinary-token KV component expected when
four branch copies are replaced by one, because total peak memory also
contains model weights, branch-specific prefixes, activations, and the
temporary branch-major view used by attention.

\section{Additional Ablations}

\subsection{Aggregation Ablations}

\begin{table*}[t]
\centering
\fontsize{8}{10}\selectfont
\setlength{\tabcolsep}{1.1mm}
\renewcommand{\arraystretch}{1.12}
\resizebox{\linewidth}{!}{
\begin{tabular}{l|ccccccccc|cccc}
\toprule
\multirow{2}{*}{\textbf{Aggregation}}
 & \multicolumn{1}{c}{\textbf{General}} & \multicolumn{4}{c}{\textbf{Math}}
 & \multicolumn{4}{c|}{\textbf{OCR}} & \multicolumn{4}{c}{\textbf{Average}} \\
\cmidrule(lr){2-2} \cmidrule(lr){3-6} \cmidrule(lr){7-10} \cmidrule(lr){11-14}
 & MMMU$_{\text{dev-val}}$ & MathVista$_{\text{mini}}$ & MathVision$_{\text{mini}}$ & LogicVista & WeMath
 & ChartQA$_{\text{test}}$ & TextVQA$_{\text{val}}$ & DocVQA$_{\text{val}}$ & OCRBench
 & General & Math & OCR & \textbf{All} \\
\midrule
Mean & 40.4 & 45.5 & 21.1 & 27.5 & 11.4
& 72.0 & 72.0 & 82.6 & 74.9
& \textbf{40.4} & 26.4 & 75.4 & 49.7 \\
MLP & 39.7 & 45.0 & 22.4 & 30.0 & 11.3
& 72.9 & 71.5 & 82.2 & 75.5
& 39.7 & \textbf{27.2} & \textbf{75.5} & \textbf{50.1} \\
\bottomrule
\end{tabular}
}
\caption{\textbf{Per-benchmark comparison of mean and MLP aggregation.} The variants are trained separately under $P_v = 4, P_l = 1$ using the same training and evaluation protocols, with all branches active at inference. MLP is used in the main experiments, and All is the equal-weight average of the nine displayed benchmarks.}
\label{tab:aggregation_comparison}
\end{table*}

Following the dynamic weighted aggregation design of ParScale~\citep{chen2025parallel}, the MLP aggregator concatenates token-aligned branch states, predicts softmax-normalized weights, and optionally smooths them toward a uniform distribution. Vision and language use the same formulation with independent parameters. Table~\ref{tab:aggregation_comparison} compares separately trained MLP and mean-aggregation variants under $P_v = 4, P_l = 1$. Both variants use the same data, optimization recipe, and evaluation protocol.

For the controlled comparison under $P_v = 4, P_l = 1$, MLP improves Math (27.2 vs.~26.4), OCR (75.5 vs.~75.4), and the overall average (50.1 vs.~49.7), while mean aggregation performs better on General (40.4 vs.~39.7). The modest 0.4-point overall gain and the domain-level trade-off indicate a small observed benefit from learned token-wise weighting in this vision-heavy setting.

\subsection{Empirical Preferences of Fixed Branches}

\begin{table*}[t]
\centering
\setlength{\tabcolsep}{4.5pt}
\small
\resizebox{\textwidth}{!}{%
\begin{tabular}{ll|ccccccccc|cccc}
\toprule
\multirow{2}{*}{\textbf{Component}}
& \multirow{2}{*}{\textbf{Branch ID}}
& \multicolumn{1}{c}{\textbf{General}}
& \multicolumn{4}{c}{\textbf{Math}}
& \multicolumn{4}{c|}{\textbf{OCR}}
& \multicolumn{4}{c}{\textbf{Average}} \\
\cmidrule(lr){3-3} \cmidrule(lr){4-7} \cmidrule(lr){8-11} \cmidrule(lr){12-15}
&
& MMMU$_{\text{dev-val}}$
& MathVista$_{\text{mini}}$ & MathVision$_{\text{mini}}$ & LogicVista & WeMath
& ChartQA$_{\text{test}}$ & TextVQA$_{\text{val}}$ & DocVQA$_{\text{val}}$ & OCRBench
& General & Math & OCR & All \\
\midrule
ViT & 0 & 39.1 & 45.9 & 21.6 & 29.5 & 11.9 & 71.3 & 71.4 & 82.1 & 76.6 & 39.1 & 27.2 & 75.4 & 49.9 \\
ViT & 1 & 38.9 & 46.0 & 20.7 & 29.0 & 12.1 & 71.3 & 71.5 & 82.1 & 76.2 & 38.9 & 27.0 & 75.3 & 49.7 \\
ViT & 2 & 38.6 & 45.9 & 20.9 & 30.0 & 12.0 & 71.3 & 71.4 & 82.0 & 76.3 & 38.6 & 27.2 & 75.3 & 49.8 \\
ViT & 3 & 38.9 & 45.6 & 20.7 & 28.4 & 12.2 & 71.4 & 71.5 & 82.0 & 76.4 & 38.9 & 26.7 & 75.4 & 49.7 \\
\midrule
LLM & 0 & 38.0 & 43.4 & 20.9 & 27.4 & 11.5 & 71.9 & 70.8 & 81.2 & 76.9 & 38.0 & 25.8 & 75.2 & 49.1 \\
LLM & 1 & 39.4 & 46.7 & 21.0 & 30.5 & 11.7 & 71.9 & 71.6 & 82.5 & 77.0 & 39.4 & 27.5 & 75.8 & 50.3 \\
LLM & 2 & 39.2 & 46.2 & 19.3 & 30.7 & 12.1 & 71.6 & 71.8 & 82.6 & 76.8 & 39.2 & 27.1 & 75.7 & 50.0 \\
LLM & 3 & 38.8 & 47.0 & 22.7 & 28.3 & 13.0 & 69.9 & 71.6 & 81.8 & 74.8 & 38.8 & 27.7 & 74.5 & 49.8 \\
\bottomrule
\end{tabular}%
}

\vspace{3pt}

\resizebox{\textwidth}{!}{%
\begin{tabular}{l|ccccccccc|cccc}
\toprule
\multirow{2}{*}{\textbf{Selection}}
& \multicolumn{1}{c}{\textbf{General}}
& \multicolumn{4}{c}{\textbf{Math}}
& \multicolumn{4}{c|}{\textbf{OCR}}
& \multicolumn{4}{c}{\textbf{Average}} \\
\cmidrule(lr){2-2} \cmidrule(lr){3-6} \cmidrule(lr){7-10} \cmidrule(lr){11-14}
& MMMU$_{\text{dev-val}}$
& MathVista$_{\text{mini}}$ & MathVision$_{\text{mini}}$ & LogicVista & WeMath
& ChartQA$_{\text{test}}$ & TextVQA$_{\text{val}}$ & DocVQA$_{\text{val}}$ & OCRBench
& General & Math & OCR & All \\
\midrule
\textbf{ViT--LLM pair}
& $(0,2)$, $(3,1)$
& $(3,3)$ & $(0,3)$, $(3,3)$ & $(2,2)$ & $(3,3)$
& $(1,0)$ & $(1,2)$ & $(1,2)$ & $(0,2)$
& $(0,2)$, $(3,1)$ & $(3,3)$ & $(0,2)$ & $(0,1)$ \\
\textbf{Score}
& 39.9
& 47.7 & 23.7 & 32.0 & 13.5
& 72.1 & 71.8 & 82.7 & 77.2
& 39.9 & 28.1 & 75.8 & 50.4 \\
\bottomrule
\end{tabular}%
}
\caption{\textbf{Fixed-branch performance from a jointly trained $4{:}4$ ParVL checkpoint.} The upper block averages each branch over its four possible partners, while the lower block reports post-hoc best pairs as oracle diagnostics. All is the equal-weight average of the nine displayed benchmarks.}
\label{tab:fixed_branch_specialization}
\end{table*}

To test whether jointly trained branches behave identically, we evaluate all 16 fixed pairs formed by one of four ViT branches and one of four LLM branches from the same checkpoint trained under $P_v = 4, P_l = 4$. Table~\ref{tab:fixed_branch_specialization} reports both marginal scores, obtained by averaging each branch over its four possible partners, and post-hoc best pairs from the full $4\times4$ branch-pair matrix. On General, the LLM-branch marginal averages range from 38.0 for branch 0 to 39.4 for branch 1, while the ViT-branch averages range from 38.6 to 39.1. The OCR differences are concentrated on the language side: LLM branches 1 and 2 reach 75.8 and 75.7 on average, whereas branch 3 reaches 74.5. The four ViT branches differ by only 0.1 points on OCR. On Math, LLM branch 0 has the lowest marginal average (25.8), whereas branches 1 and 3 reach 27.5 and 27.7, respectively.

The post-hoc best pair varies across benchmarks and domain averages, with no pair consistently dominating. Because pair selection uses the same evaluation scores reported in the table, these results are oracle diagnostics rather than a deployable routing policy or a fair performance comparison. They reveal checkpoint-specific performance differences among branches, not stable semantic specialization across tasks within a domain. The branches receive no expert labels or domain-specific losses, and branch IDs can be permuted without changing the architecture. We therefore do not interpret a numbered branch as an intrinsic OCR or Math expert. Establishing stable semantic specialization would require replication across training seeds and direct representation or routing analyses.

\section{Factorized Sparse Router}
\label{sec:appendix_sparse_router}

The learned-router result reported in the main paper uses the jointly trained ParVL checkpoint with $P_v = 4$ and $P_l = 4$. It selects one ViT--LLM branch pair for each input and executes only that pair. The resulting model therefore replaces dense $4{:}4$ aggregation with sparse per-sample inference.

\paragraph{Sequential routing.}
The router has two small 4-way MLP gates. The first gate reads a pooled image thumbnail feature together with frozen text embeddings and selects a visual branch $v\in\{0,1,2,3\}$. We run only this visual branch and pass its connector feature $h_v$ to the second gate. The second gate reads $h_v$ and the text embeddings, then selects a language branch $l\in\{0,1,2,3\}$. Generation uses only $(v,l)$. This follows the top-1 sparse-execution principle of sparse MoE routing~\citep{fedus2022switch}, while factorizing the choice into a visual gate followed by a connector-conditioned language gate. The factorization avoids a separate 16-way joint gate while preserving the dependency of the language choice on the selected visual representation.

\paragraph{Offline path teacher.}
We first freeze the $P_v = 4, P_l = 4$ ParVL checkpoint at training step 7{,}000 and compute the token-level negative log-likelihood (NLL) of the ground-truth answer for all 16 possible branch pairs on each router-training sample. Let $c_{v,l}$ denote the NLL for pair $(v,l)$; lower values indicate a better path for that sample. These 16 values are stored before router training. They are not recomputed during router training.

We convert the 16 NLLs into a soft teacher distribution, assigning higher probability to lower-NLL paths. The ViT teacher is the marginal probability of each visual branch after summing over its four language branches. The LLM teacher is the four-way distribution over language branches conditional on the ViT branch currently selected by the ViT router. Thus, the first gate learns ``which visual branch is promising for this input,'' while the second learns ``given that visual representation, which language branch should answer.''

More concretely, with teacher temperature $\tau=0.1$, the path teacher is
\begin{equation}
q(v,l\mid x)=\operatorname*{softmax}_{(v,l)}\left(-\frac{c_{v,l}-\min_{v',l'}c_{v',l'}}{\tau}\right),
\end{equation}
where the softmax is normalized over all 16 pairs and the subtracted minimum is a shift included for numerical stability.
The visual target is $q_v(v\mid x)=\sum_l q(v,l\mid x)$. If the current visual gate selects $\hat v$, the language target is $q_l(l\mid \hat v,x)\propto\exp(-c_{\hat v,l}/\tau)$. Writing $p_v$ and $p_l$ for the softmax outputs of the two gates and $\operatorname{CE}(q,p)=-\sum q\log p$, the router loss averages two cross-entropies:
\begin{equation}
\mathcal{L}_{\mathrm{router}}=\tfrac{1}{2}\operatorname{CE}(q_v,p_v)+\tfrac{1}{2}\operatorname{CE}(q_l,p_l).
\end{equation}
For example, if the offline labels show that $v_1$ is consistently favorable and $(v_1,l_2)$ has the lowest NLL, the first loss increases $p_v(v_1)$ and, once $v_1$ is selected, the second loss increases $p_l(l_2)$. The targets remain soft, so near-optimal alternatives retain probability mass.

\paragraph{Training and inference.}
We freeze the ViT, LLM, connector, prefixes, and aggregation modules, and train only the two router gates. The gates use hidden dimension 256 and are trained with 16 GPUs for 1200 optimizer steps, global batch size 128, learning rate $3\times10^{-4}$, cosine decay, and 3\% warmup. The per-device batch size is one, and gradients are accumulated for eight steps, giving $16\times1\times8=128$ samples per update. This is supervised path-level distillation~\citep{hinton2015distilling}, not REINFORCE or straight-through optimization of an online generation reward.

At inference, both gates use argmax top-1 selection. The router checkpoint is loaded alongside the frozen model trained under $P_v = 4, P_l = 4$, and the active-branch mechanism skips the remaining visual and language branches. This is therefore a computation-saving routing policy rather than output-level reweighting.

Practical deployment would additionally require latency evaluation under
realistic request mixtures, router calibration, and validation across
additional checkpoints and training distributions.

\end{document}